\documentclass[journal]{IEEEtai}

\usepackage[colorlinks,urlcolor=blue,linkcolor=blue,citecolor=blue]{hyperref}

\usepackage{color,array}
\usepackage[export]{adjustbox}
\usepackage{graphicx}
\usepackage{times}
\usepackage{multirow}
\makeatletter
\DeclareRobustCommand\onedot{\futurelet\@let@token\@onedot}
\def\@onedot{\ifx\@let@token.\else.\null\fi\xspace}
\def\eg{\emph{e.g}\onedot} \def\Eg{\emph{E.g}\onedot}
\def\ie{\emph{i.e}\onedot} \def\Ie{\emph{I.e}\onedot}
\makeatother
\usepackage{amssymb}
\usepackage{subfig}
\usepackage{pgfplots}
\pgfplotsset{compat=1.14}
\usepgfplotslibrary{groupplots}
\usetikzlibrary{patterns}

\newcommand{\mymethod}{RAPiD}
\usepackage{amsmath,amsfonts}
\usepackage{algorithmic}
\usepackage{algorithm}
\usepackage{array}
\usepackage{textcomp}
\usepackage{stfloats}
\usepackage{url}
\usepackage{verbatim}
\usepackage{graphicx}
\usepackage{cite}
\usepackage{times}
\usepackage{soul}
\usepackage{url}
 \usepackage[utf8]{inputenc}
\usepackage[font=small]{caption} \usepackage{amsthm}
\usepackage{xspace}
\usepackage{xcolor}
\usepackage{booktabs}
\usepackage{algorithm}
\usepackage{algorithmic}
\usepackage[switch]{lineno}
\usepackage{multirow}

\usepackage{tikz}
\usetikzlibrary{positioning, fit, backgrounds}

\makeatletter
\DeclareRobustCommand\onedot{\futurelet\@let@token\@onedot}
\def\@onedot{\ifx\@let@token.\else.\null\fi\xspace}
\def\eg{\emph{e.g}\onedot} \def\Eg{\emph{E.g}\onedot}
\def\ie{\emph{i.e}\onedot} \def\Ie{\emph{I.e}\onedot}
\makeatother
\usepackage{amssymb}
\usepackage{pgfplots}
\pgfplotsset{compat=1.14}
\usepgfplotslibrary{groupplots}
\usetikzlibrary{patterns}

\begin{document}

% \title{RAPiD: Real-time Deterministic Trajectory Planning via Diffusion Behavior Priors for Safe and Efficient Autonomous Driving} 

\title{RAPiD: Reward-Guided Consistency Distillation of Diffusion Planners for Real-Time Autonomous Driving}

\author{
Ruturaj Reddy$^{1,2}$, Hrishav Bakul Barua$^{1,3*}$, Junn Yong Loo$^{5*}$, Thanh Thi Nguyen$^{2,4}$, Ganesh Krishnasamy$^1$\\
$^1$School of Information Technology, Monash University, Malaysia \\
$^2$Faculty of Information Technology, Monash University, Australia \\
$^3$TCS Research, India \\
$^4$School of Science, Technology and Engineering, University of the Sunshine Coast, Australia \\
$^5$School of Electrical and Electronic Engineering, Nanyang Technological University, Singapore \\

{ruturaj.reddy, hrishav.barua, ganesh.krishnasamy}@monash.edu, junnyong.loo@ntu.edu.sg, tnguyen5@usc.edu.au \\
$^*$equal contributions
}

\markboth{}
{Reddy \MakeLowercase{\textit{et al.}}}

\maketitle

\begin{abstract}

    Diffusion-based trajectory planners can model multi-modal driving behavior, but their iterative denoising process introduces a latency bottleneck for real-time closed-loop deployment. We present \mymethod, a reward-guided consistency distillation framework that distills a pretrained DiffusionPlanner into a few-step consistency student while retaining multi-modal trajectory generation. The student is trained using deterministic teacher denoising steps from the frozen diffusion planner, together with a low-noise data anchor that keeps generated trajectories grounded in expert demonstrations. To make distillation safety-aware, we train an Implicit Q-Learning critic on a balanced mixture of ground-truth log-replay and DiffusionPlanner rollout trajectories, each scored using a modified PDM-style reward, providing trajectory-level supervision beyond conventional imitation learning. During deployment, the 2-step student generates $K$ trajectories, and the trained critic performs best-of-$K$ trajectory selection conditioned on the latent state. On nuPlan, \mymethod~maintains comparable performance to the diffusion teacher on non-reactive closed-loop splits and remains competitive on reactive splits, while reducing complete-pipeline inference latency from 100.91 ms to 18.41 ms, corresponding to a $5.5\times$ speedup. On interPlan, \mymethod~achieves the highest aggregate score among learning-based methods, demonstrating competitive generalization in interactive long-tail scenarios. These results show that reward-guided consistency distillation can convert a pretrained diffusion planner into a few-step closed-loop planner that substantially reduces inference cost while preserving safety-oriented trajectory selection.
    
\end{abstract}

\begin{IEEEImpStatement}
Autonomous systems must make safe and reliable decisions within strict real-time constraints, yet many state-of-the-art diffusion-based planners remain too computationally expensive for practical deployment. This work demonstrates that high-quality, multimodal planning can be achieved with substantially lower inference latency, enabling faster, smoother, and safety-oriented decision-making for autonomous vehicles. Beyond self-driving cars, the proposed approach may benefit a wide range of robotics applications, including autonomous delivery, warehouse automation, and industrial robotics, where rapid and reliable decision-making is essential. By showing that planning efficiency can be improved without sacrificing behavioral diversity or safety, this work contributes toward developing more practical, trustworthy, and deployable AI systems for safety-critical real-world applications.
\end{IEEEImpStatement}


\section{Introduction}
\begin{figure}[h]
    \centering
    \includegraphics[width=1\linewidth]{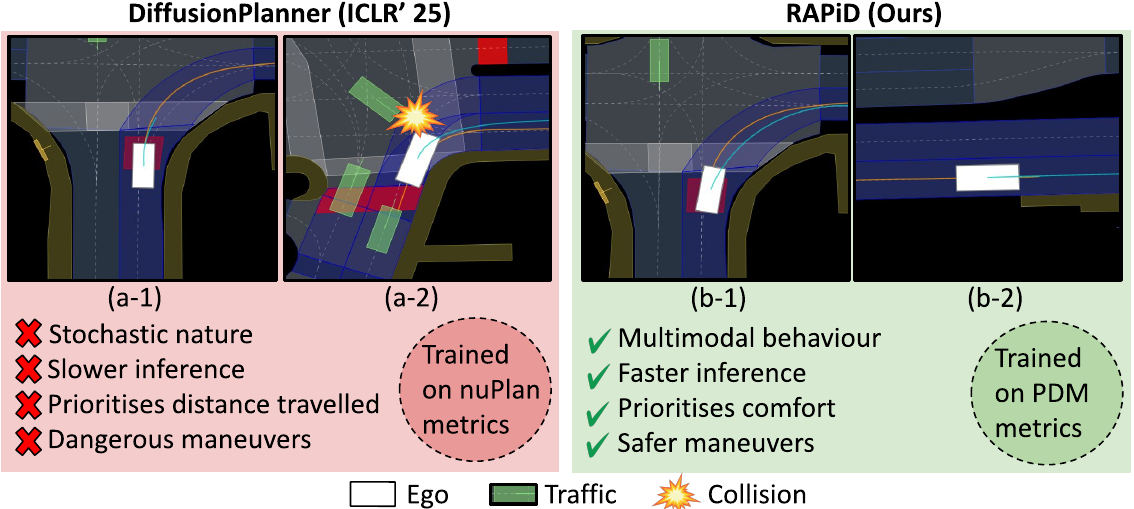}
    \caption{Figs.~(a-1) and (a-2) illustrate the limitations of the baseline DiffusionPlanner. DiffusionPlanner reacts later in this qualitative example and subsequently collides, while requiring 100.91 ms of complete-pipeline inference latency. In contrast, Figs.~(b-1) and (b-2) demonstrate our proposed RAPiD framework. By distilling the diffusion prior into a few-step multimodal policy trained on PDM safety metrics rather than nuPlan metrics, RAPiD achieves $5.5\times$ faster inference. This efficiency enables timely decisions, resulting in a smooth, collision-free maneuver.\looseness=-1}
    \label{fig:cover_image}
\end{figure}

\IEEEPARstart{I}{n} autonomous driving, generating safe and interactive trajectories in complex traffic scenarios remains a fundamental challenge~\cite{xia2024survey,hu2025survey,ganesan2024comprehensive,chen2024end}. Traditional rule-based systems provide interpretability, but they often struggle to generalize across long-tail events and highly interactive traffic situations~\cite{treiber2000congested}. As a result, planning research has increasingly shifted toward large-scale data-driven planning, including generative models that can capture multi-modal driving behavior and generate diverse trajectories~\cite{scheel2022urban,rhinehart2018deep,chen2024end,bharilya2024machine,peng2025diffusion}. Among these, score-based diffusion models~\cite{song2021scorebased}, and specifically DiffusionPlanner~\cite{zheng2025diffusion}, have achieved strong closed-loop performance on high-fidelity benchmarks such as nuPlan~\cite{karnchanachari2024towards}. Their strength lies in modeling the multimodal structure of human driving behavior. Here, multimodality refers to the existence of several distinct, physically valid future trajectory modes under the same scene context, such as merging left, yielding in-lane, or continuing forward at different speeds. Preserving these distinct trajectory modes is critical in interactive traffic, because a regression policy that averages them can produce an unsafe interpolated trajectory that commits to none of the valid maneuvers~\cite{codevilla2018end,shafiullah2022behavior}. However, this expressiveness comes with a deployment cost, \ie, diffusion planners generate trajectories through iterative denoising, requiring multiple network evaluations for trajectory generation. This creates a latency bottleneck for closed-loop autonomous driving, where planning must be both multimodal and responsive under strict real-time constraints.

Several recent approaches have attempted to reduce the diffusion-latency bottleneck, but each leaves open a different trade-off for closed-loop driving. Deterministic extraction methods such as SRPO~\cite{chen2024score} avoid iterative sampling by compiling a diffusion prior into a single fast policy, but their mode-seeking objective commits to one behavior and can discard alternative valid maneuvers. More broadly, recent surveys on diffusion models for intelligent transportation systems and knowledge distillation identify iterative sampling cost as a central barrier to deploying diffusion-based models in latency-sensitive settings~\cite{peng2025diffusion,moslemi2024survey}. Consistency models provide a technical route to few-step generation by learning mappings along the diffusion trajectory~\cite{song2023consistency,kim2024consistency,song2024improved}. However, existing consistency-based formulations do not directly address closed-loop autonomous driving, where the planner must preserve multiple valid trajectory modes while selecting among them using safety- and interaction-aware rewards. Thus, existing acceleration approaches do not directly address reward-guided distillation of a pretrained diffusion planner in which a learned safety critic guides both consistency distillation and deployment-time best-of-$K$ trajectory selection.

To address this gap, we propose \mymethod, a reward-guided consistency distillation framework that distills a pretrained DiffusionPlanner into a few-step closed-loop planner. Unlike deterministic extraction methods that reduce the diffusion prior to a single output, \mymethod~distills the frozen DiffusionPlanner teacher~\cite{zheng2025diffusion} into a discrete consistency student that preserves the multiple valid trajectory modes. This student model is trained using deterministic DDIM teacher steps, which provide adjacent denoising targets from the frozen diffusion prior without requiring the full iterative sampling chain. This allows the student to approximate the teacher's generative behavior while substantially reducing the number of network function evaluations. A low-noise data anchor is added to stabilize the high-noise consistency bootstrap and keep generated trajectories grounded in expert driving demonstrations. To make distillation safety-aware, we train an implicit Q-Learning critic~\cite{kostrikov2021offline} on the frozen latent state encodings produced by the pretrained DiffusionPlanner encoder. This critic is supervised using a balanced mixture of ground-truth log-replay and DiffusionPlanner rollout trajectories scored with the modified PDM-style reward adopted in Diffusion-ES~\cite{dauner2023parting,diffusionES}, emphasizing collision avoidance, drivable-area compliance, time-to-collision, comfort, speed-limit compliance, progress, and proximity to leading agents. During distillation, the critic is applied with a small reward weight to steer the student toward safer in-support modes without overpowering the generative prior. At deployment, the 2-step student generates multiple trajectories, and the trained critic performs best-of-$K$ trajectory selection conditioned on the latent state to choose the highest-reward trajectory for execution. Fig.~\ref{fig:cover_image} illustrates this trade-off, \ie, the diffusion baseline reacts later in a challenging closed-loop scene, whereas \mymethod~produces a smoother and timely maneuver with substantially lower inference latency.

Extensive experiments validate both the efficiency and safety-oriented behavior of the proposed framework across high-fidelity closed-loop driving benchmarks and controlled offline-RL tasks. Beyond aggregate performance, our evaluations isolate the load-bearing mechanisms of the proposed architecture and reveal how different consistency-distillation topologies behave under reward guidance. The main contributions of this work are summarized as follows:
\begin{itemize}

    \item We introduce \mymethod, a reward-guided consistency distillation framework for closed-loop autonomous driving. \mymethod~compresses a pretrained DiffusionPlanner teacher into a few-step discrete consistency student using deterministic DDIM teacher steps, while a low-noise data anchor keeps the student grounded in expert driving trajectories.

    \item We integrate a PDM-trained implicit Q-Learning critic into both training and deployment. During distillation, the critic provides a small reward-guidance signal that steers the student toward safer in-support trajectory modes. During inference, the same critic performs best-of-$K$ selection over batched 2-step candidate trajectories conditioned on the frozen latent state.

    \item We show through controlled ablations that both deployment components are load-bearing, \ie, critic ranking, rather than random candidate sampling, provides the selection gain, while the second denoising step is required to produce valid multimodal candidates. On our evaluation setup, \mymethod~reduces complete-pipeline latency from 100.91 ms to 18.41 ms, a $5.5\times$ speedup over the diffusion teacher.

    \item We evaluate \mymethod~on nuPlan, interPlan, and D4RL. The method matches or slightly improves the diffusion teacher on non-reactive nuPlan splits, remains competitive on reactive splits with improved safety-oriented PDM submetrics, and demonstrates strong generalization among learning-based planners on interPlan. In D4RL, we further identify a reward-guidance dichotomy: discrete-time students maintain mode coverage in multimodal navigation, whereas continuous-time consistency students collapse under reward guidance in the same regime.

\end{itemize}

\section{Related Work}
\subsection{Diffusion and Generative Trajectory Planners}

Recent surveys of autonomous-driving planning and end-to-end driving systems highlight the shift from modular rule-based pipelines toward learning-based planners that jointly model perception, prediction, and planning~\cite{ganesan2024comprehensive, chen2024review, reda2024path}. Traditional autonomous-driving planners rely on rule-based, search-based, or optimization-based decision logic, including car-following models, lane-graph search, model predictive control, and hand-designed safety constraints~\cite{treiber2000congested,dauner2023parting}. These methods are interpretable and often reliable in structured scenarios, but they require substantial engineering effort to handle rare, interactive, and long-tail traffic situations. Learning-based planners reduce this manual design burden by learning directly from driving logs, with imitation-learning and transformer-based approaches showing strong performance on closed-loop benchmarks~\cite{scheel2022urban,huang2023gameformer,cheng2024rethinking,cheng2024pluto}. However, deterministic or regression-based policies remain vulnerable when the same scene admits multiple distinct valid futures. In such cases, minimizing an averaged imitation loss can produce a trajectory between valid modes, which may be neither physically plausible nor safe.

Generative trajectory models address this limitation by modeling a distribution over possible futures rather than a single point estimate~\cite{rhinehart2018deep,janner2022planning,ajayconditional,chi2025diffusion}. Diffusion models are especially attractive because their iterative denoising process can represent complex multimodal behavior distributions while conditioning on map, route, ego-history, and surrounding-agent context~\cite{ho2020denoising,song2021scorebased}. In autonomous driving, DiffusionPlanner~\cite{zheng2025diffusion} demonstrates that a transformer-based diffusion planner can jointly model ego planning and neighboring-agent prediction, achieving strong closed-loop performance without rule-based refinement. DiffusionES~\cite{diffusionES} further combines diffusion priors with gradient-free trajectory search to optimize black-box driving rewards. These works establish diffusion as a powerful policy class for planning, but they inherit the deployment cost of iterative sampling, \ie, high-quality generation requires repeated network evaluations, and maintaining diversity often requires sampling multiple candidate trajectories. Our work targets this efficiency bottleneck while retaining the multimodal behavior distribution learned by a pretrained diffusion planner.

\begin{figure*}[h]
    \centering
    \includegraphics[width=1.0\linewidth]{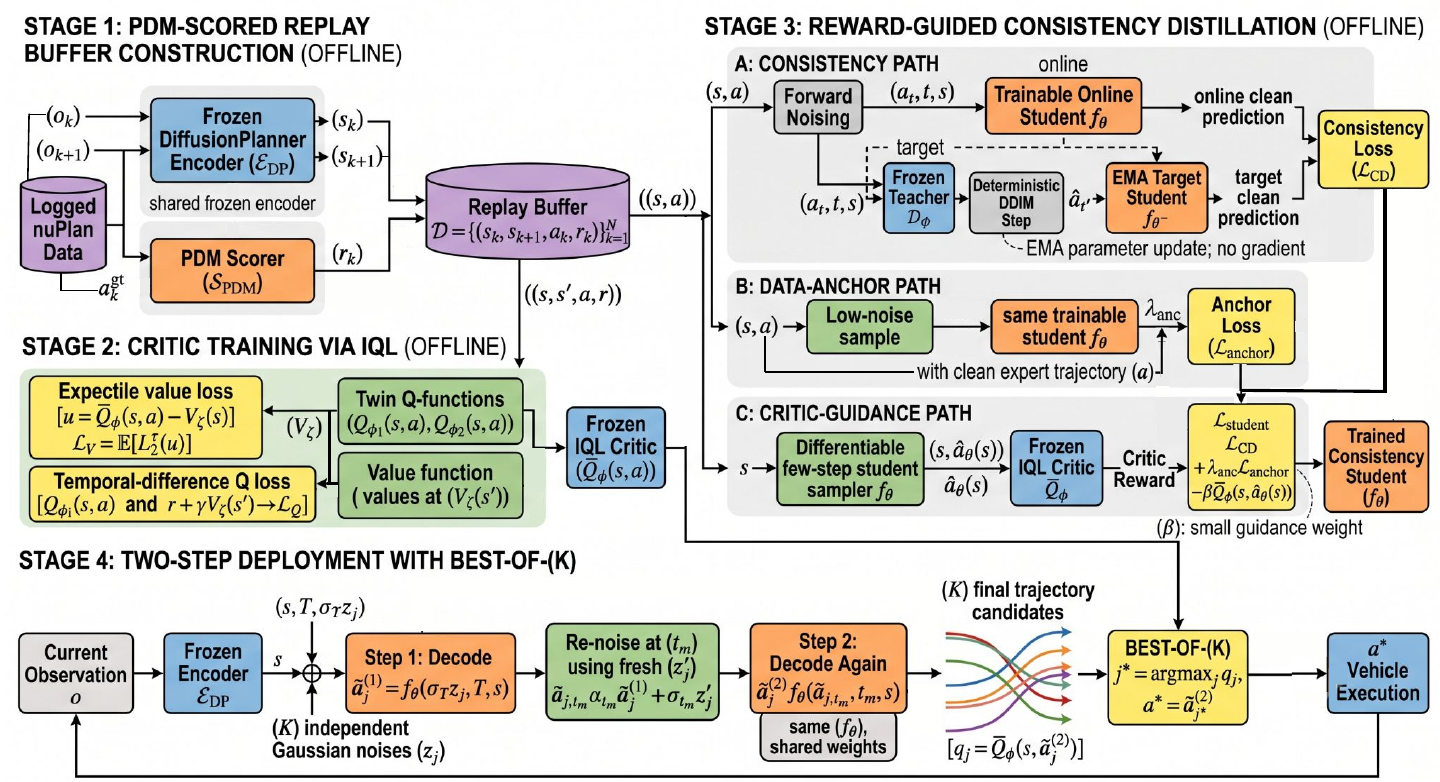}
    \caption{\textbf{Overview of the proposed \mymethod~framework} (Section~\ref{sec:method}). \textit{Stage 1} constructs the offline replay buffer from logged nuPlan data. Current and next driving contexts are encoded by the shared frozen DiffusionPlanner~\cite{zheng2025diffusion} encoder to obtain $s_k$ and $s_{k+1}$, while each ground-truth log-replay or DiffusionPlanner rollout trajectory $a_k$ is evaluated by the PDM scorer to obtain the safety and comfort-oriented reward $r_k$, forming replay tuples $(s,s',a,r)$. \textit{Stage 2} trains an Implicit Q-Learning critic using expectile regression for $V_\zeta$ and temporal-difference learning for the twin Q-functions, while the resulting averaged critic $\bar Q_\phi$ is then frozen. \textit{Stage 3} distills the frozen DiffusionPlanner teacher into a few-step consistency student through three complementary paths, \ie, deterministic DDIM teacher targets and an EMA target student provide consistency supervision, a low-noise data anchor grounds predictions in expert trajectories, and a small frozen-critic guidance term steers generated trajectories toward higher-value in-support modes. \textit{Stage 4} performs online two-step deployment. For each of $K$ Gaussian initializations, the trained student first decodes a preliminary trajectory, re-noises it at an intermediate time using fresh noise, and decodes it again to obtain $K$ final candidates. The frozen critic scores these candidates, best-of-$K$ selection chooses the highest-valued trajectory $a^\star$ for execution, and the procedure repeats in a receding-horizon manner.}
    
    \label{fig:framework_overview}
\end{figure*}

\subsection{Accelerating and Distilling Generative Policies}

A broad line of work accelerates diffusion models by improving the reverse sampling process. Higher-order solvers such as DDIM and DPM-Solver reduce the number of denoising steps required at inference~\cite{song2020denoising,lu2022dpm}, but they still rely on an iterative reverse process. Consistency models provide a more direct alternative by learning to map noisy samples to clean data along the diffusion trajectory, enabling one-step or few-step generation~\cite{song2023consistency}. Consistency Trajectory Models generalize this idea by learning mappings between arbitrary points along the probability-flow trajectory~\cite{kim2024consistency}, while improved consistency training introduces more stable training procedures and robust losses~\cite{song2024improved}. These methods provide the technical foundation for few-step generative modeling, but they do not by themselves determine how a planner should choose among several valid trajectory modes when safety and interaction rewards differ.

In offline reinforcement learning and decision-making, generative policies are often coupled with value functions to select high-reward actions. Diffuser and related planning-as-generation methods use diffusion models for trajectory-level decision-making~\cite{janner2022planning}, while Diffusion-QL and IDQL use diffusion policies to represent multimodal action distributions and combine them with critic-based selection~\cite{wang2023diffusion,hansen2023idql}. These methods preserve expressive action distributions, but inference still requires diffusion sampling. SRPO~\cite{chen2024score} avoids this cost by extracting a fast deterministic policy from a pretrained diffusion prior using score regularization. However, its mode-seeking objective collapses the prior to a single committed behavior at inference. Reward-aware consistency trajectory distillation~\cite{duan2025accelerating} shows that reward information can be incorporated into consistency distillation for offline RL. In contrast, our work studies reward-aware consistency distillation in closed-loop autonomous driving, where the student must simultaneously preserve valid trajectory modes, remain close to a pretrained diffusion planner, and support safety-guided selection under real-time constraints.

\subsection{Safety-Aware Offline RL and Closed-Loop Driving Evaluation}

Offline reinforcement learning is a natural framework for safety-aware planning because it allows logged demonstrations to be reweighted or selected according to reward signals rather than only imitation likelihood. Methods such as IQL~\cite{kostrikov2021offline} learn value functions from static datasets without requiring explicit behavior-policy estimation, reducing extrapolation to out-of-distribution actions. Advantage-weighted regression and related behavior-regularized methods use learned values to emphasize higher-quality actions while remaining close to the data distribution~\cite{peng2020advantage,kumar2019stabilizing,kostrikov2021offline}. However, in driving, the reward must reflect closed-loop safety and interaction quality, not merely proximity to expert demonstrations. This motivates critics trained on trajectory-level safety scores, which can rank multiple plausible candidates according to collision risk, time-to-collision, comfort, drivable-area compliance, and progress.

Closed-loop benchmarks such as nuPlan~\cite{karnchanachari2024towards} and interPlan~\cite{hallgarten2024can} expose the limitations of open-loop displacement metrics by evaluating planners inside interactive simulation. Prior work has shown that open-loop accuracy can be misaligned with closed-loop driving quality, and that rule-based or hybrid planners such as PDM can remain competitive by enforcing safety-relevant priors~\cite{dauner2023parting}. The PDM scorer aggregates safety and quality terms such as collision avoidance, drivable-area compliance, driving direction, time-to-collision, comfort, and progress, making it a useful signal for distinguishing safe from unsafe trajectory candidates~\cite{dauner2023parting,diffusionES}. Our method uses this perspective not only for evaluation but also for learning, \ie, PDM-scored ground-truth log-replay and DiffusionPlanner rollout trajectories supervise an IQL critic, which then guides consistency distillation and performs deployment-time best-of-K trajectory selection.

Despite these advances, existing approaches do not directly address reward-guided distillation of pretrained diffusion planners for closed-loop autonomous driving. \mymethod~fills this gap by distilling a frozen DiffusionPlanner teacher into a 2-step discrete consistency student, while a PDM-trained critic guides distillation and selects among candidate trajectories during deployment.

\section{Methodology}
\subsection{Problem Formulation}

We consider closed-loop trajectory planning for an ego vehicle in an interactive driving scene. At decision step $k$, the planner observes a raw scene context $o_k$, which includes ego history, surrounding-agent states, map elements, traffic-light states, route information, and static scene context available in the nuPlan setting~\cite{karnchanachari2024towards}. The objective is to predict a future ego trajectory over a planning horizon of $H$ steps,
\begin{equation}
    a_k = \{(x_{k+i}, y_{k+i}, \psi_{k+i})\}_{i=1}^{H},
\end{equation}
where $(x,y)$ denotes position and $\psi$ denotes heading. Throughout this paper, we use $a_k$ to denote the action in the offline-RL formulation. This action corresponds to the full future trajectory rather than a low-level control command.

To connect trajectory planning with offline reinforcement learning, each logged transition is represented by a current context $o_k$, next context $o_{k+1}$, trajectory $a_k$ obtained from either ground-truth log replay or a DiffusionPlanner rollout, and scalar reward $r_k$. The current and next latent states are obtained using the frozen DiffusionPlanner encoder,
\begin{equation}
    s_k = \mathcal{E}_{\mathrm{DP}}(o_k),
    \qquad
    s_{k+1} = \mathcal{E}_{\mathrm{DP}}(o_{k+1}),
\end{equation}
where $\mathcal{E}_{\mathrm{DP}}$ remains fixed throughout training. The reward is computed by evaluating each replay trajectory—ground-truth log replay or DiffusionPlanner rollout—with the PDM scorer.
\begin{equation}
    r_k = \mathcal{S}_{\mathrm{PDM}}(o_k, a_k).
\end{equation}
The resulting offline dataset consists of replay tuples
\begin{equation}
    \mathcal{D}
    =
    \left\{
    (s_k, s_{k+1}, a_k, r_k)
    \right\}_{k=1}^{N}.
\end{equation}
For readability, we denote each tuple as $(s,s',a,r)$ in the remainder of the paper.

\subsection{Pretrained Diffusion Planner}

We use a pretrained DiffusionPlanner~\cite{zheng2025diffusion} as the frozen generative teacher. DiffusionPlanner is built on a diffusion-transformer architecture~\cite{peebles2023scalable} and jointly models ego planning and neighboring-agent prediction, allowing it to represent multiple valid future trajectory modes under the same scene context. In \mymethod, we reuse both its encoder and denoising prior, \ie, the encoder produces latent scene embeddings used by the critic and student, while the denoising prior provides teacher targets for consistency distillation. All DiffusionPlanner parameters remain frozen.

Let $a$ denote a clean expert trajectory and let $s=\mathcal{E}_{\mathrm{DP}}(o)$ denote the frozen latent state embedding. The forward diffusion process corrupts $a$ at diffusion time $t\in[0,1]$ according to
\begin{equation}
    q_{t|0}(a_t|a)
    =
    \mathcal{N}(a_t \mid \alpha_t a,\sigma_t^2 I),
\end{equation}
or equivalently,
\begin{equation}
    a_t = \alpha_t a + \sigma_t \epsilon,
    \qquad
    \epsilon \sim \mathcal{N}(0,I),
\end{equation}
where $\alpha_t$ and $\sigma_t$ are the teacher noise-schedule coefficients. We denote the frozen DiffusionPlanner denoiser by $D_\psi$. Given a noised trajectory $a_t$, diffusion time $t$, and latent state $s$, the teacher predicts a clean trajectory estimate
\begin{equation}
    \hat a_0^\psi = D_\psi(a_t,t,s).
\end{equation}
If the implementation is parameterized as noise prediction $\epsilon_\psi(a_t,t,s)$, the corresponding clean prediction is obtained as
\begin{equation}
    \hat a_0^\psi
    =
    \frac{a_t-\sigma_t\epsilon_\psi(a_t,t,s)}{\alpha_t}.
\end{equation}

This pretrained diffusion prior provides an expressive multimodal trajectory distribution, but sampling from it requires an iterative denoising process with multiple network function evaluations. We therefore do not deploy the teacher directly. Instead, we use deterministic DDIM teacher steps derived from this frozen denoiser as supervision for the few-step consistency student introduced in Section~\ref{sec:method}.

\subsection{Consistency Distillation for Generative Policies}

A direct way to reduce diffusion-policy inference cost is to avoid sampling from the diffusion model at deployment. Score Regularized Policy Optimization (SRPO)~\cite{chen2024score} follows this idea by extracting a deterministic policy from a pretrained diffusion behavior prior. Its objective can be written as
\begin{align}
\begin{split}
    \max_\theta \mathcal{L}_\pi(\theta)
    =
    \mathbb{E}_{s\sim\mathcal{D},\,a\sim\pi_\theta}
    \left[ Q_\phi(s,a) \right]
    -
    \frac{1}{\beta}
    D_{\mathrm{KL}}
    \left[
    \pi_\theta(\cdot|s)
    \,\|\, 
    \mu(\cdot|s)
    \right],
\end{split}
\end{align}
where $\mu(\cdot|s)$ denotes the pretrained behavior prior and $Q_\phi$ is a learned critic. The reverse-KL term makes the extracted policy mode-seeking: it encourages the policy to select a high-value mode of the behavior distribution rather than cover all plausible modes. This property is useful for efficient deterministic control, but it is undesirable in interactive driving scenes where multiple distinct maneuvers may be valid under the same context. A single extracted action can discard alternative safe modes that would be useful during closed-loop interaction.

Consistency models~\cite{song2023consistency} provide a different route to fast inference. Instead of extracting a single deterministic action, a consistency student learns to map noised samples back to clean outputs along the teacher denoising trajectory, enabling one-step or few-step generation. Extensions such as Consistency Trajectory Models~\cite{kim2024consistency} and improved consistency training~\cite{song2024improved} further stabilize few-step generation by improving the trajectory-matching objective and training losses. In decision-making, reward-aware consistency trajectory distillation~\cite{duan2025accelerating} shows that value guidance can steer a consistency student toward higher-reward samples. These methods suggest a promising alternative to deterministic extraction, \ie, compress the diffusion prior into a few-step generative student while preserving the ability to sample multiple valid modes.

In \mymethod, we adopt this consistency-distillation perspective for closed-loop trajectory planning. Rather than using SRPO to extract a single deterministic policy, we distill the frozen DiffusionPlanner teacher into a discrete consistency student that can generate multiple candidate trajectories. The critic is retained, but its role changes, \ie, it provides a small reward-guidance signal during distillation and ranks sampled candidates during deployment. This design preserves the multimodal structure of the diffusion teacher while avoiding the full iterative diffusion sampling chain.

\subsection{Safety-Aware Reward Design via PDM Scorer}

The critic requires a trajectory-level reward that reflects closed-loop driving quality rather than only imitation likelihood. We therefore use the modified PDM-Closed scoring function adopted in Diffusion-ES~\cite{diffusionES} as the trajectory-level reward for critic supervision. Given a raw scene context $o_k$ and replay trajectory $a_k$, obtained from either ground-truth log replay or a DiffusionPlanner rollout, the reward is defined as
\begin{equation}
    r_k = \mathcal{S}_{\mathrm{PDM}}(o_k, a_k).
\end{equation}
The PDM score aggregates safety- and comfort-oriented criteria, including collision avoidance, time-to-collision (TTC), drivable-area compliance, driving-direction compliance, speed-limit compliance, progress, passenger comfort, and proximity to the leading agent.

We use this PDM reward for critic supervision, not as a replacement for the official closed-loop benchmark scores. The purpose is to train a critic that can distinguish safer and more comfortable trajectory candidates within the support of the logged driving data. In particular, the enhanced PDM scorer used in Diffusion-ES~\cite{diffusionES} includes additional penalties for unsafe proximity to leading agents and speed-limit violations, which encourages the critic to prefer trajectories with larger safety margins and smoother behavior. The resulting critic is later used both during reward-guided consistency distillation and during deployment-time best-of-$K$ trajectory selection. Additional details of the PDM metrics are provided in Section S1 of the \textit{Supplementary Material}.

\subsection{Proposed Method}
\label{sec:method}

Fig.~\ref{fig:framework_overview} summarizes the proposed \mymethod~pipeline. The method consists of three stages: (i) construction of a PDM-scored offline replay buffer using the frozen DiffusionPlanner encoder, (ii) Implicit Q-Learning critic training on the scored latent-state transitions, and (iii) reward-guided consistency distillation of the frozen DiffusionPlanner teacher into a few-step student. At deployment, the student generates a batch of candidate trajectories using a 2-step sampler, and the trained critic selects the highest-reward candidate.

\textbf{Stage 1: PDM-Scored Replay Buffer Construction.}
We construct the offline replay buffer from a balanced combination of transitions collected through ground-truth log replay and DiffusionPlanner simulation~\cite{hester2018deep,ball2023efficient}.
For each recorded nuPlan simulation transition, let $o_k$ denote the raw driving context at planning step $k$, including ego history, surrounding-agent states, map information, traffic-light states, and route information. At each step, $a_k$ denotes the complete 8-second future ego trajectory obtained from ground-truth log replay or predicted by DiffusionPlanner. The simulator advances to the next planning step using this trajectory, producing the next context $o_{k+1}$. We use the frozen DiffusionPlanner encoder $\mathcal{E}_{\mathrm{DP}}$ to obtain
\begin{equation}
    s_k = \mathcal{E}_{\mathrm{DP}}(o_k),
    \qquad
    s_{k+1} = \mathcal{E}_{\mathrm{DP}}(o_{k+1}).
\end{equation}
The trajectory $a_k$ is treated as the offline-RL action rather than a low-level control command. Each trajectory is evaluated using the PDM scorer:
\begin{equation}
    r_k = \mathcal{S}_{\mathrm{PDM}}(o_k,a_k).
\end{equation}
The reward reflects collision avoidance, drivable-area compliance, driving-direction compliance, time-to-collision, speed-limit compliance, progress, comfort, and proximity to the leading agent. The resulting offline replay buffer is
\begin{equation}
    \mathcal{D}
    =
    \left\{
    \left(s_k,s_{k+1},a_k,r_k\right)
    \right\}_{k=1}^{N}.
\end{equation}
For readability, we write each tuple as $(s,s',a,r)$ in the remainder of the paper.

\textbf{Stage 2: Critic Training via Implicit Q-Learning.}
We train an IQL critic on the PDM-scored replay buffer. The critic consists of a value function $V_\zeta(s)$ and twin action-value functions $Q_{\phi_1}(s,a)$ and $Q_{\phi_2}(s,a)$, where $a$ is a candidate future trajectory. We define the averaged critic as
\begin{equation}
    \bar Q_\phi(s,a)
    =
    \frac{1}{2}
    \left(
    Q_{\phi_1}(s,a) + Q_{\phi_2}(s,a)
    \right),
\end{equation}
where $\phi=\{\phi_1,\phi_2\}$. The value function is trained using expectile regression:
\begin{equation}
    \mathcal{L}_V(\zeta)
    =
    \mathbb{E}_{(s,a)\sim\mathcal{D}}
    \left[
    L_2^\tau
    \left(
    \bar Q_\phi(s,a)-V_\zeta(s)
    \right)
    \right],
\end{equation}
where
\begin{equation}
    L_2^\tau(u)
    =
    \left|\tau-\mathbf{1}_{\{u<0\}}\right|u^2,
    \qquad \tau \in (0.5,1).
\end{equation}
Using $\tau>0.5$ biases the value estimate toward higher-return trajectories while avoiding explicit behavior-policy queries. The twin Q-functions are trained with the temporal-difference objective
\begin{equation}
    \mathcal{L}_Q(\phi)
    =
    \mathbb{E}_{(s,s',a,r)\sim\mathcal{D}}
    \left[
    \sum_{i=1}^{2}
    \left\|
    r+\gamma V_\zeta(s')-Q_{\phi_i}(s,a)
    \right\|_2^2
    \right],
\end{equation}
where $\gamma$ is the discount factor. After training, the critic is frozen and the averaged value $\bar Q_\phi(s,a)$ is used both as a reward-guidance signal during consistency distillation and as the ranking function for deployment-time best-of-$K$ selection.

\textbf{Stage 3: Reward-Guided Consistency Distillation.}
We distill the frozen DiffusionPlanner teacher into a few-step consistency student
$f_\theta(a_t,t,s)$, which maps a noised trajectory $a_t$ at diffusion time $t$ back to a clean trajectory conditioned on the latent state $s$. Given a clean expert trajectory $a$, forward noising is defined as
\begin{equation}
    a_t = \alpha_t a + \sigma_t \epsilon,
    \qquad
    \epsilon \sim \mathcal{N}(0,I),
\end{equation}
where $\alpha_t$ and $\sigma_t$ follow the frozen teacher's noise schedule.

For an adjacent pair of time steps $t'<t$, the frozen DiffusionPlanner teacher provides a deterministic DDIM denoising step from $a_t$ to an earlier noised sample $\hat a_{t'}$. Writing the teacher clean-trajectory prediction as $\hat a_0^\psi = D_\psi(a_t,t,s)$, the implied noise estimate is
\begin{equation}
    \hat\epsilon_\psi
    =
    \frac{a_t-\alpha_t \hat a_0^\psi}{\sigma_t},
\end{equation}
and the corresponding DDIM teacher target is
\begin{equation}
    \hat a_{t'}
    =
    \alpha_{t'}\hat a_0^\psi
    +
    \sigma_{t'}\hat\epsilon_\psi.
\end{equation}
This teacher step is held under stop-gradient. With an exponential-moving-average target network $f_{\theta^-}$, the consistency loss is
\begin{equation}
    \mathcal{L}_{\mathrm{CD}}(\theta)
    =
    \mathbb{E}_{t,\epsilon}
    \left[
    \omega(t)\,
    d\!\left(
    f_\theta(a_t,t,s),
    f_{\theta^-}(\hat a_{t'},t',s)
    \right)
    \right],
\end{equation}
where $\omega(t)$ is a time-dependent weighting term.

To stabilize the high-noise consistency bootstrap, we add a low-noise data anchor. For $t_d \sim \mathcal{U}(0,t_{\max}^{\mathrm{anc}})$ and
$a_{t_d}=\alpha_{t_d}a+\sigma_{t_d}\epsilon$, the anchor loss is
\begin{equation}
    \mathcal{L}_{\mathrm{anchor}}(\theta)
    =
    \mathbb{E}_{t_d,\epsilon}
    \left[
    d\!\left(
    f_\theta(a_{t_d},t_d,s),a
    \right)
    \right].
\end{equation}
Here, $d$ denotes the per-dimension Huber loss with $\delta=1$, summed over the trajectory dimensions. Finally, the frozen critic provides a small reward-guidance term. Let $\hat a_\theta(s)=\mathrm{clip}\!\left( f_\theta(\sigma_T z,T,s),\pm1\right)$, where $z\sim\mathcal{N}(0,I)$, denote the differentiable one-step readout used for critic guidance. The complete student objective is
\begin{equation}
\begin{split}
    \mathcal{L}_{\mathrm{student}}(\theta)
    =
    \mathcal{L}_{\mathrm{CD}}(\theta)
    +
    \lambda_{\mathrm{anc}}\mathcal{L}_{\mathrm{anchor}}(\theta)
    -
    \beta
    \mathbb{E}_{s}
    \left[
    \bar Q_\phi(s,\hat a_\theta(s))
    \right],
\end{split}
\end{equation}
where $\lambda_{\mathrm{anc}}$ controls the data-anchor strength and $\beta$ is kept small so that critic guidance steers the student among in-support modes without overpowering the frozen diffusion prior.

\textbf{Deployment: 2-Step Sampling with Critic Best-of-$K$.}
At inference, the raw scene context is encoded once as $s=\mathcal{E}_{\mathrm{DP}}(o)$. The student draws $K$ candidate trajectories using a 2-step sampler. Each candidate is initialized from Gaussian noise, decoded once at the highest noise level, re-noised to a lower intermediate time, and decoded again:
\begin{equation}
    \tilde a_k^{(1)} = f_\theta(\sigma_T z_k,T,s),
    \qquad
    z_k \sim \mathcal{N}(0,I),
\end{equation}
\begin{equation}
    \tilde a_{k,t_m}
    =
    \alpha_{t_m}\tilde a_k^{(1)}
    +
    \sigma_{t_m}z'_k,
    \qquad
    z'_k \sim \mathcal{N}(0,I),
\end{equation}
\begin{equation}
    \tilde a_k^{(2)}
    =
    f_\theta(\tilde a_{k,t_m},t_m,s).
\end{equation}
The final executed trajectory is selected by the frozen critic:
\begin{equation}
    j^\star = \arg\max_{j\in\{1,\dots,K\}} \bar Q_\phi(s,\tilde a_j^{(2)}),
    \qquad
    a^\star=\tilde a_{j^\star}^{(2)}.
\end{equation}
Thus, multimodality is preserved by candidate generation, while safety-oriented trajectory selection is performed by the PDM-trained critic.

\section{Experiments}
\subsection{Experiment Setup}

\mymethod~is implemented in PyTorch~\cite{paszke2019pytorch} and trained using an NVIDIA RTX 4090 GPU with 24GB VRAM. The main driving evaluation is conducted on nuPlan~\cite{karnchanachari2024towards} and interPlan~\cite{hallgarten2024can}. We report nuPlan scores for closed-loop planning quality, interPlan scores for scenario-specific interactive performance, and PDM-style scorer metrics~\cite{diffusionES} for safety and comfort analysis. Representative comparison methods are listed in Tables~\ref{tab:nuPlan_benchmark} and~\ref{tab:interPlan_scores}. Additional controlled offline-RL experiments on D4RL~\cite{fu2020d4rl}, including AntMaze and Gym-MuJoCo tasks, are provided in S4 of the \textit{Supplementary Material}. For those supplementary experiments, we reuse SRPO pretrained diffusion and critic checkpoints as the base initialization to isolate the effect of the proposed consistency-distillation and critic-selection modifications under a controlled comparison.

\subsection{Evaluation Results}
\textbf{nuPlan Benchmark Performance.} 
We evaluate our method against SOTA baselines, including learning-based planners (PDM-Open~\cite{dauner2023parting}, UrbanDriver~\cite{scheel2022urban}, Gameformer~\cite{huang2023gameformer}, Plan-TF~\cite{chi2025diffusion}, PLUTO~\cite{cheng2024pluto}, DiffusionPlanner~\cite{zheng2025diffusion}) and rule-based planners (IDM~\cite{treiber2000congested}, PDM-Closed~\cite{dauner2023parting}, PDM-Hybrid~\cite{dauner2023parting}). Table~\ref{tab:nuPlan_benchmark} summarizes performance on the nuPlan benchmark. On non-reactive (NR) splits, \mymethod~maintains comparable performance to the DiffusionPlanner baseline while achieving marginally higher scores on Val14, Test14, and Test14-hard. Specifically, \mymethod~achieves NR scores of 90.01, 89.88, and 76.01, compared with 89.87, 89.19, and 75.99 for DiffusionPlanner. Table~\ref{tab:inference_time} reports complete-pipeline inference latency, including input feeding, normalization, encoding, and output generation. \mymethod~reduces latency from 100.91 ms for DiffusionPlanner to 18.41 ms, corresponding to a $5.5\times$ speedup, and is also substantially faster than UrbanDriver. \mymethod~is slower than lightweight single-output baselines such as PDM-Open, PlanTF, and PLUTO because it generates multiple trajectories with a 2-step consistency sampler and ranks them using the critic for best-of-$K$ trajectory selection. This reflects the intended trade-off of \mymethod: retaining multimodal trajectory generation and safety-oriented selection while substantially reducing the iterative sampling cost of the diffusion teacher.

\begin{table*}[t]
    \centering
    % --- Left Table: nuPlan Benchmark ---
    \begin{minipage}{0.72\textwidth}
        \centering
        \caption{Closed-loop planning results on nuPlan benchmark. Bold values indicate best performance in each category. *: Methods using pre-computed reference paths benefit from additional prior information. \textbf{NR}: non-reactive evaluation. \textbf{R}: reactive evaluation.}
        \label{tab:nuPlan_benchmark}
        \resizebox{\linewidth}{!}{%
            \begin{tabular}{llcccccc}
            \toprule
            \multirow{2}{*}{\textbf{Type}} & \multirow{2}{*}{\textbf{Methods}} & \multicolumn{2}{c}{\textbf{Val14}} & \multicolumn{2}{c}{\textbf{Test14-hard}} & \multicolumn{2}{c}{\textbf{Test14}} \\
            \cmidrule(lr){3-4} \cmidrule(lr){5-6} \cmidrule(lr){7-8}
            & & NR & R & NR & R & NR & R \\
            \midrule
            \textcolor{gray}{Expert} & \textcolor{gray}{Log-replay} & \textcolor{gray}{93.53} & \textcolor{gray}{80.32} & \textcolor{gray}{85.96} & \textcolor{gray}{68.80} & \textcolor{gray}{94.03} & \textcolor{gray}{75.86} \\
            \midrule
            \multirow{6}{*}{\shortstack{Rule-based \& \\Hybrid}}
            & IDM~\cite{treiber2000congested} & 75.60 & 77.33 & 56.15 & 62.26 & 70.39 & 74.42 \\
            & PDM-Closed~\cite{dauner2023parting} & 92.84 & 92.12 & 65.08 & 75.19 & 90.05 & 91.63 \\
            & PDM-Hybrid~\cite{dauner2023parting} & 92.77 & 92.11 & 65.99 & 76.07 & 90.10 & 91.28 \\
            & GameFormer~\cite{huang2023gameformer} & 79.94 & 79.78 & 68.70 & 67.05 & 83.88 & 82.05 \\
            & PLUTO~\cite{cheng2024pluto} & 92.88 & 76.88 & \textbf{80.08} & 76.88 & 92.23 & 90.29 \\
            & Diffusion Planner w/ refine~\cite{zheng2025diffusion} & \textbf{94.26} & \textbf{92.90} & 78.87 & \textbf{82.00} & \textbf{94.80} & \textbf{91.75} \\
            \midrule
            \multirow{6}{*}{Learning-based}
            & PDM-Open*~\cite{dauner2023parting} & 53.53 & 54.24 & 33.51 & 35.83 & 52.81 & 57.23 \\
            & UrbanDriver~\cite{scheel2022urban} & 68.57 & 64.11 & 50.40 & 49.95 & 51.83 & 67.15 \\
            & GameFormer w/o refine~\cite{huang2023gameformer} & 13.32 & 8.69 & 7.08 & 6.69 & 11.36 & 9.31 \\
            & PlanTF~\cite{chi2025diffusion} & 84.27 & 76.95 & 69.70 & 61.61 & 85.62 & 79.58 \\
            & PLUTO w/o refine*~\cite{cheng2024pluto} & 88.89 & 78.11 & 70.03 & 59.74 & \textbf{89.90} & 78.62 \\
            & DiffusionPlanner~\cite{zheng2025diffusion} & 89.87 & \textbf{82.80} & 75.99 & \textbf{69.22} & 89.19 & \textbf{82.93} \\
            & \textbf{RAPiD (Ours)} & \textbf{90.01} & \underline{82.13} & \textbf{76.01} & \underline{67.34} & 89.88 & \underline{82.01} \\
            \bottomrule
            \end{tabular}%
        }
    \end{minipage}%
    \hfill % Adds space between the minipages
    % --- Right Table: Inference Time ---
    \begin{minipage}{0.25\textwidth}
        \centering
        \caption{Inference time comparison across learning-based methods. Note: Reported times measure the complete pipeline (input feeding, normalization, encoding, and output generation), rather than output generation alone.}
        \label{tab:inference_time}
        \resizebox{\linewidth}{!}{%
            \begin{tabular}{lc}
            \toprule
            \textbf{Method} & \textbf{Time (ms)} \\
            \midrule
            PDM-Open~\cite{dauner2023parting} & 14.67 \\
            UrbanDriver~\cite{scheel2022urban} & 78.42 \\
            PlanTF~\cite{chi2025diffusion} & \textbf{4.51} \\
            Pluto~\cite{cheng2024pluto} & 9.19 \\
            DiffusionPlanner~\cite{zheng2025diffusion} & 100.91 \\
            \midrule
            \textbf{\mymethod~(Ours)} & 18.41 \\
            \bottomrule
            \end{tabular}%
        }
    \end{minipage}
\end{table*}

At deployment, the consistency student generates $K$ trajectories using a 2-step sampler, and the averaged twin critic selects the final trajectory through best-of-$K$ trajectory selection. Fig.~\ref{fig:bestofk_sweep} sweeps $K$ on Val14 reactive evaluation. Closed-loop score improves with diminishing returns and saturates at $K{=}32$, with no further gain at larger $K$. This behavior suggests that increasing $K$ improves coverage of high-quality trajectory modes up to a point, after which additional trajectories provide limited benefit and may expose the selection process to critic overestimation. We therefore use $K{=}32$ in deployment. Since the $K$ trajectories are generated in a single batched pass, best-of-$K$ selection adds limited overhead relative to the 2-step student forward pass. This sweep serves as a deployment ablation for the best-of-$K$ trajectory selection used in nuPlan. Since all other inference settings are fixed, the improvement from $K{=}1$ to $K{=}32$ indicates that generating multiple trajectories and selecting with the critic contributes to the closed-loop gain, while the saturation beyond $K{=}32$ motivates our deployment choice.

\begin{figure*}[h]
    \centering
    
    % ==========================================
    % ROW 1: Four Square Plots in ONE Line
    % ==========================================
    \subfloat{
        \begin{minipage}{0.45\columnwidth}
            \centering
            \includegraphics[width=\linewidth,frame]{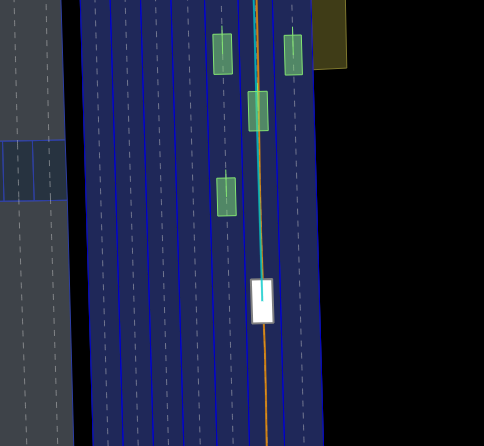}\\[0.5ex]
            \tiny (a-1) DiffusionPlanner
        \end{minipage}
    }
    \subfloat{
        \begin{minipage}{0.45\columnwidth}
            \centering
            \includegraphics[width=\linewidth,frame]{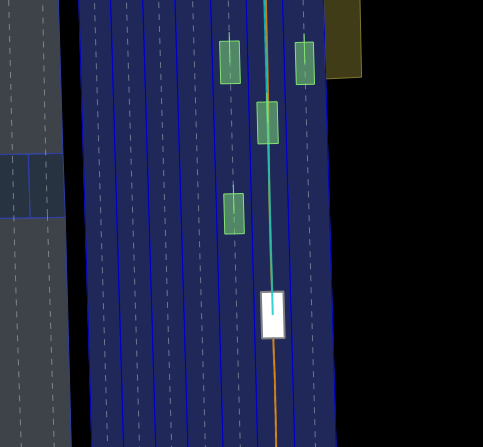}\\[0.5ex]
            \tiny (a-2) RAPiD
        \end{minipage}
    }
    \subfloat{
        \begin{minipage}{0.45\columnwidth}
            \centering
            \includegraphics[width=\linewidth,frame]{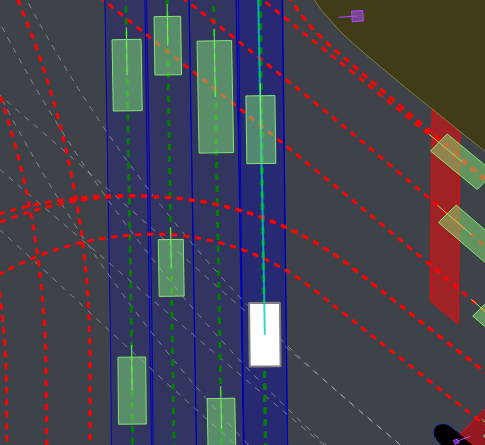}\\[0.5ex]
            \tiny (b-1) DiffusionPlanner
        \end{minipage}
    }
    \subfloat{
        \begin{minipage}{0.45\columnwidth}
            \centering
            \includegraphics[width=\linewidth,frame]{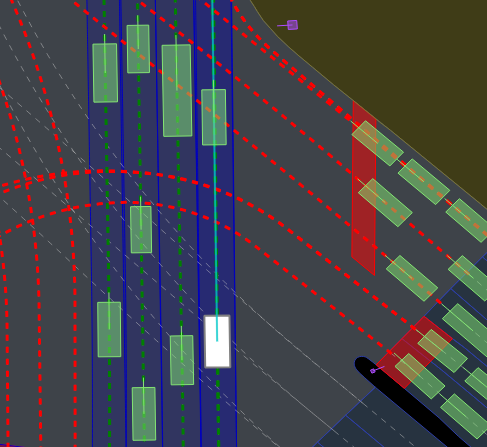}\\[0.5ex]
            \tiny (b-2) RAPiD
        \end{minipage}
    }

    \vspace{0.2em}

    % ==========================================
    % ROW 2: Two Rectangular Metric Plots
    % ==========================================
    \subfloat{
        \begin{minipage}{0.9\columnwidth}
            \centering
            \includegraphics[width=\linewidth,frame]{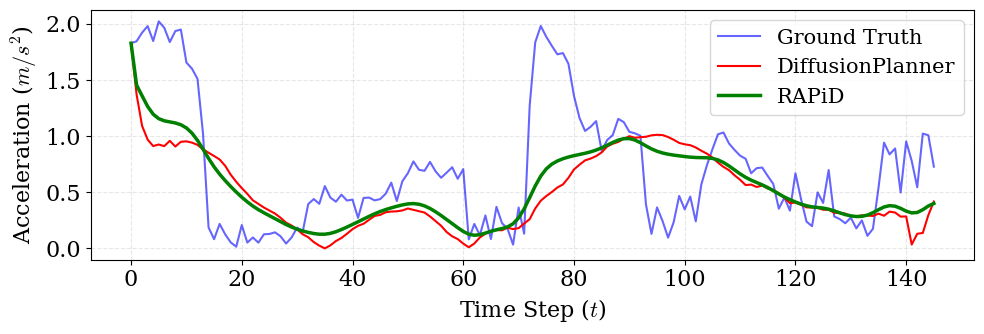}\\[0.5ex]
            \tiny Acceleration Metrics
            \label{acceleration_metric}
        \end{minipage}
    }
    \subfloat{
        \begin{minipage}{0.95\columnwidth}
            \centering
            \includegraphics[width=\linewidth,frame]{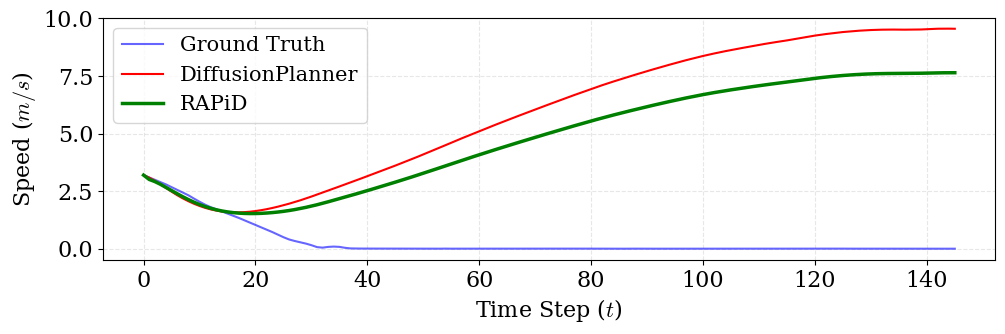}\\[0.5ex]
            \tiny Speed Metrics
            \label{speed_metric}
        \end{minipage}
    }
    
    \vspace{0.2em}
    
    % ==========================================
    % ROW 3: Four Square Plots in ONE Line
    % ==========================================
    \subfloat{
        \begin{minipage}{0.45\columnwidth}
            \centering
            \includegraphics[width=\linewidth,frame]{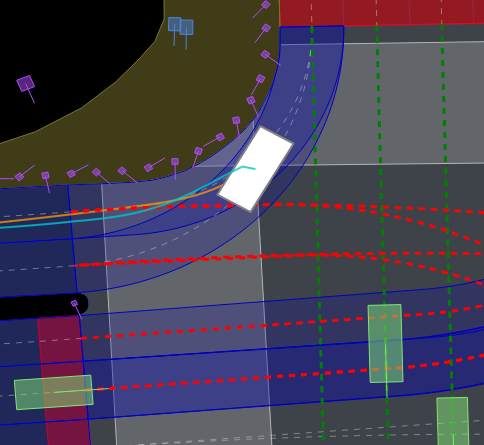}\\[0.5ex]
            \tiny (c-1) DiffusionPlanner
        \end{minipage}
    }
    \subfloat{
        \begin{minipage}{0.45\columnwidth}
            \centering
            \includegraphics[width=\linewidth,frame]{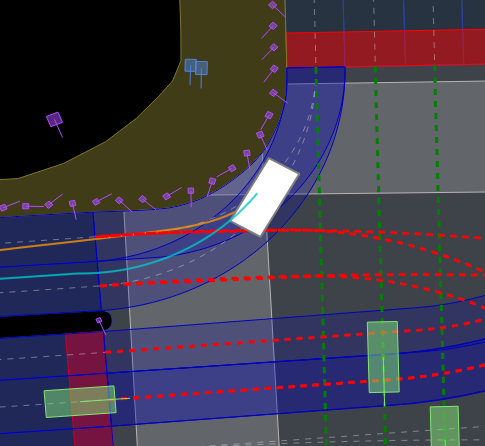}\\[0.5ex]
            \tiny (c-2) RAPiD
        \end{minipage}
    }
    \subfloat{
        \begin{minipage}{0.45\columnwidth}
            \centering
            \includegraphics[width=\linewidth,frame]{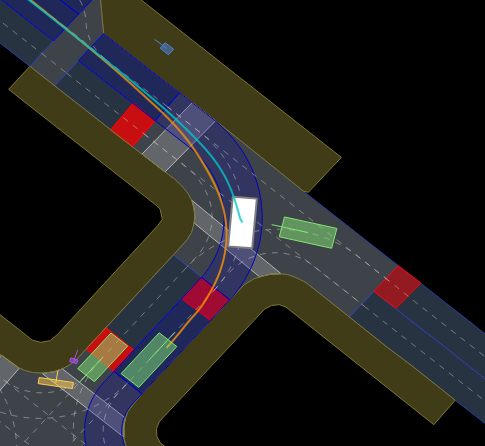}\\[0.5ex]
            \tiny (d-1) DiffusionPlanner
        \end{minipage}
    }
    \subfloat{
        \begin{minipage}{0.45\columnwidth}
            \centering
            \includegraphics[width=\linewidth,frame]{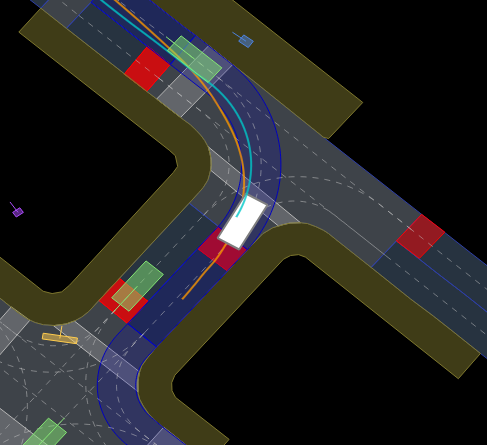}\\[0.5ex]
            \tiny (d-2) RAPiD
        \end{minipage}
    }

    \vspace{0.2em}

    % ==========================================
    % ROW 4: Two Rectangular Metric Plots
    % ==========================================
    \subfloat{
        \begin{minipage}{0.95\columnwidth}
            \centering
            \includegraphics[width=\linewidth,frame]{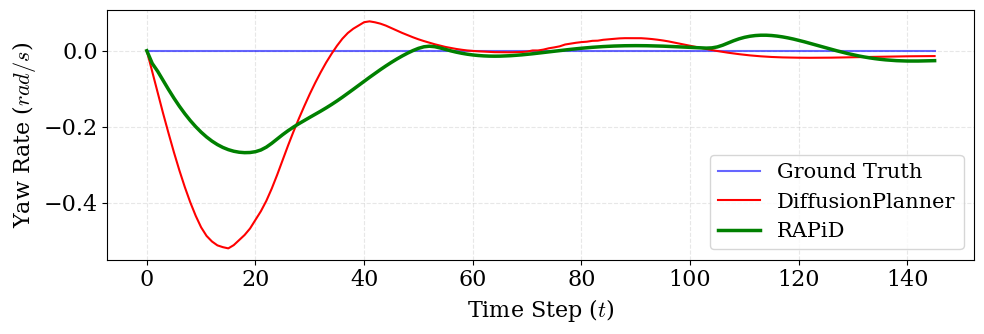}\\[0.5ex]
            \tiny Yaw Rate Metrics
            \label{yaw_rate_1}
        \end{minipage}
    }
    \subfloat{
        \begin{minipage}{0.95\columnwidth}
            \centering
            \includegraphics[width=\linewidth,frame]{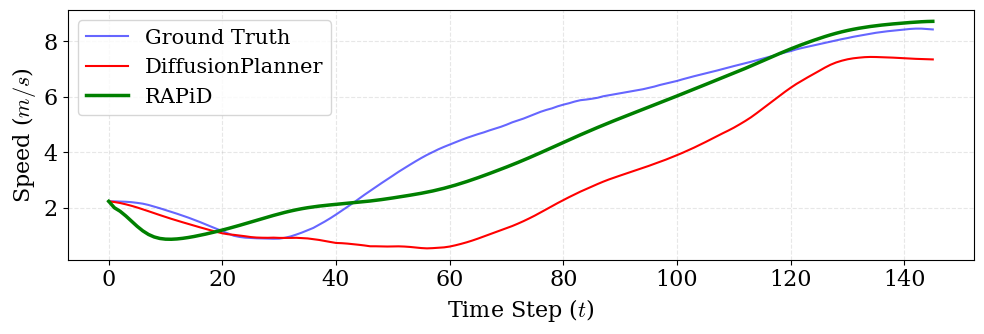}\\[0.5ex]
            \tiny Speed Metrics
            \label{yaw_rate_2}
        \end{minipage}
    }
    
    \caption{Qualitative results across four scenarios in nuPlan: Following Lane, Stopping with Lead, Starting Right Turn, and Low Speed Maneuvering. Each figure displays the generated trajectories for DiffusionPlanner and RAPiD (top) along with their corresponding Comfort metrics (bottom) including acceleration, speed, and yaw rate demonstrating the smoother control exerted by RAPiD compared to the baseline DiffusionPlanner.}
    \label{fig:qualitative_results}
\end{figure*}

In the reactive (R) environment, \mymethod~achieves competitive performance, obtaining 82.01 on Test14 compared with 82.93 for DiffusionPlanner. The reactive gap reflects a safety-progress trade-off induced by the PDM-style reward, which places stronger emphasis on comfort, proximity, and safety margins while reducing the relative emphasis on progress compared with the standard nuPlan scorer. In lane-following scenarios, \mymethod~maintains closer alignment with the lane centerline and produces smoother acceleration patterns when following lead vehicles (Fig.~\ref{fig:qualitative_results}(a-2)), compared with the DiffusionPlanner baseline in the corresponding scene (Fig.~\ref{fig:qualitative_results}(a-1)). During stopping scenarios, \mymethod~reduces speed earlier and maintains larger following distances (Fig.~\ref{fig:qualitative_results}(b-2)), whereas DiffusionPlanner exhibits shorter following distances in the corresponding example (Fig.~\ref{fig:qualitative_results}(b-1)). For lane changes, \mymethod~executes more gradual merges with lower yaw rates (Fig.~\ref{fig:qualitative_results}(c-2)) compared with the sharper maneuvers generated by DiffusionPlanner (Fig.~\ref{fig:qualitative_results}(c-1)).

At intersections, DiffusionPlanner can exhibit hesitation followed by abrupt maneuvers, leading to near-collision or collision-prone behavior associated with delayed maneuvers (Fig.~\ref{fig:qualitative_results}(d-1); Fig.~\ref{fig:cover_image}(a-2)). In contrast, \mymethod~produces smoother and more timely turns with controlled acceleration (Fig.~\ref{yaw_rate_2}(d-2)) and avoids the collision-prone behavior shown in the corresponding baseline examples (Fig.~\ref{fig:cover_image}(b-1,b-2)). See Section S7-A in the \textit{Supplementary Material} for a detailed frame-by-frame analysis. These qualitative differences are consistent with the PDM-style reward used for critic training, which encourages trajectories with improved comfort, proximity, and safety margins, even when this reduces progress-oriented scoring in the reactive nuPlan setting.

\begin{figure}[t]
    \centering
    \definecolor{customGreen}{RGB}{141,198,141}
    \begin{tikzpicture}
        \begin{axis}[
            width=\columnwidth,
            height=5.2cm,
            xlabel={Number of candidates $K$},
            ylabel={Closed-loop Score (CLS)},
            xlabel style={font=\small},
            ylabel style={font=\small, yshift=-1mm},
            xmode=log,
            log basis x=2,
            xtick={1,4,8,16,32,64,128},
            xticklabels={1,4,8,16,32,64,128},
            xticklabel style={font=\footnotesize},
            yticklabel style={font=\footnotesize},
            ymin=77.5, ymax=82.9,
            grid=major,
            grid style={dashed, gray!25},
            enlarge x limits=0.08,
            mark options={solid},
        ]
            \addplot[
                color=customGreen!70!black,
                line width=1pt,
                mark=*, mark size=2pt,
                mark options={fill=customGreen}
            ] coordinates {
                (1,79.2) (4,80.2) (8,80.9) (16,81.6)
                (32,82.13) (64,82.0) (128,81.8)
            };
            \addplot[only marks, mark=*, mark size=3.2pt,
                     color=red!70!black, fill=red!60!white]
                coordinates {(32,82.13)};
            \node[font=\scriptsize, anchor=south west, xshift=0.5mm, yshift=0.5mm]
                at (axis cs:32,82.13) {$K{=}32$};
            \draw[dashed, red!50, line width=0.6pt]
                (axis cs:32,77.5) -- (axis cs:32,82.13);
        \end{axis}
    \end{tikzpicture}
    \caption{Closed-loop score (CLS) on nuPlan val14 (reactive) versus best-of-$K$ candidates. CLS improves with diminishing returns and saturates at $K{=}32$ (deployment setting), with no further gain at larger $K$, consistent with critic overestimation under max-selection.}
    \label{fig:bestofk_sweep}
\end{figure}

\begin{table}[t]
    \centering
    \caption{Effect of the number of function evaluations (NFE).}
    \label{tab:nfe_ablation}
    \small
    \begin{tabular}{@{}rrrrrr@{}}
        \toprule
        NFE & CLS & Progress & No Collision & TTC & Decoder Latency \\
        \midrule
        1 & 78.00 & 0.850 & 0.890 & 0.890 & 6.10 ms \\
        2 & \textbf{82.13} & \textbf{0.885} & \textbf{0.955} &
            \textbf{0.925} & 11.91 ms \\
        4 & 82.01 & 0.880 & 0.954 & 0.923 & 23.10 ms \\
        8 & 81.90 & 0.881 & 0.952 & 0.922 & 45.31 ms \\
        \bottomrule
    \end{tabular}
\end{table}

\begin{table}[t]
    \centering
    \caption{Effect of critic-based candidate selection.}
    \label{tab:selection_ablation}
    \small
    \begin{tabular}{@{}rllrrrr@{}}
        \toprule
        $K$ & Selection & CLS & Progress & No Collision & TTC \\
        \midrule
        1  & None   & 79.20 & 0.872 & 0.943 & 0.910 \\
        32 & Random & 79.15 & 0.879 & 0.939 & 0.919 \\
        32 & Critic & \textbf{82.13} & \textbf{0.885} &
             \textbf{0.955} & \textbf{0.925} \\
        \bottomrule
    \end{tabular}
\end{table}

\begin{table*}[h]
    \centering
    \caption{\textbf{Quantitative results}. The interPlan benchmark scores. interPlan is aggregated scores, followed by specific scenario type scores. \textbf{Const.}: Construction, \textbf{Acc.}: Accident, \textbf{Jayw.}: Jaywalking Pedestrians, \textbf{Overt.}: Overtaking, \textbf{LTD}: Low Traffic Density, \textbf{MTD}: Medium Traffic Density, \textbf{HTD}: High Traffic Density.}
    \label{tab:interPlan_scores}
    \scriptsize
    \begin{tabular}{l l c c c c c c c c c c}
        \toprule
        \textbf{Type} & \textbf{Method} & \textbf{interPlan} & \textbf{Const.} & \textbf{Acc.} & \textbf{Jayw.} & \textbf{Nudge} & \textbf{Overt.} & \textbf{LTD} & \textbf{MTD} & \textbf{HTD} \\
        \midrule
        \multirow{3}{*}{\shortstack{Rule-\\based}} 
        & IDM ~\cite{treiber2000congested}              & 31 & 0  & 0  & \textbf{66} & 0  & 0 & 61 & 61 & 61 \\
        & IDM+MOBIL ~\cite{kesting2007general}        & 31 & \textbf{21} & 0  & \textbf{66} & 0  & 0 & \textbf{71} & 21 & \textbf{70} \\
        & PDM-Closed ~\cite{dauner2023parting}       & \textbf{42} & 18 & 0  & 48 & \textbf{74} & \textbf{9} & 62 & \textbf{62} & 62 \\
        \midrule
        \multirow{5}{*}{\shortstack{Learning-\\based}} 
        & Urban Driver ~\cite{scheel2022urban}     & 4  & 0  & 0  & 0  & 0  & 0 & 29 & 0  & 30 \\
        & GC-PGP ~\cite{hallgarten2023prediction}           & 10 & 0  & 0  & 0  & 0  & 0 & 18 & 16 & 44 \\
        & GameFormer ~\cite{huang2023gameformer}        & 11 & 0  & 0  & 48 & 0  & 0 & 0  & 20 & 21 \\
        & PDM-Open ~\cite{dauner2023parting}          & 25 & 13 & 0  & \textbf{56} & 36 & 8 & 29 & 29 & 26 \\
        & DTPP ~\cite{huang2024dtpp}             & 25 & \textbf{18} & \textbf{18} & 44 & 10 & 0 & 40 & \textbf{36} & 34 \\
        & DiffusionPlanner ~\cite{zheng2025diffusion} & 29 & 9 & 0 & 26 & 60 & 25 & 40 & 21 & 48 \\
        & \textbf{\mymethod~(Ours)}         & \textbf{30} & 11 & 0 & 27 & \textbf{61} & \textbf{26} & \textbf{42} & 27 & \textbf{49} \\
        \bottomrule
    \end{tabular}
\end{table*}

\begin{figure}[t]
    \centering
    % Colors extracted from your screenshot (Lines)
    \definecolor{customRed}{RGB}{255,131,131}
    \definecolor{customGreen}{RGB}{141, 198, 141}

    \begin{tikzpicture}
        \begin{groupplot}[
            group style={
                group name=myplots,
                group size=1 by 3,
                vertical sep=0.5cm,
            },
            ybar,
            width=\columnwidth,
            height=3.8cm,
            ymin=0.85, ymax=1.05,
            ylabel={Score},
            ylabel style={font=\scriptsize, yshift=-1mm},
            yticklabel style={font=\tiny},
            symbolic x coords={No Col., Driv. Area, Direction, Speed Lim., Progress, TTC, Comfort, Lane Foll., Proximity},
            xtick=data,
            xticklabel style={font=\fontsize{6}{7}\selectfont, rotate=45, anchor=east},
            grid=major,
            grid style={dashed, gray!20},
            enlarge x limits=0.08,
        ]

        % ---------------------------------------------------------
        % Plot (a): Val14
        % ---------------------------------------------------------
        \nextgroupplot[
            title={(a) Val14}, 
            title style={font=\scriptsize, at={(0.5,0.9)}},
            legend style={
                at={(0.98,0.98)}, 
                anchor=north east,
                legend columns=2,
                font=\tiny,
                fill=white, 
                fill opacity=0.9, 
                text opacity=1,
                draw=black!30,
                rounded corners=2pt,
                /tikz/every even column/.append style={column sep=0.2cm}
            },
            legend image code/.code={
                \draw[#1] (0cm,-0.1cm) rectangle (0.15cm,0.1cm);
            },              
            xticklabels=\empty
        ]
            % BASELINE -> customRed solid
            \addplot[
                fill=customRed,
                draw=none,
                bar width=4pt, 
                bar shift=-2.5pt
            ] coordinates {
                (No Col., 0.9467) (Driv. Area, 1.0000) (Direction, 0.9841)
                (Speed Lim., 0.9803) (Progress, 0.9374) (TTC, 0.9116)
                (Comfort, 0.9787) (Lane Foll., 0.9237) (Proximity, 0.9694)
            };
            
            % OURS -> customGreen solid
            \addplot[
                fill=customGreen,
                draw=none,
                bar width=4pt, 
                bar shift=2.5pt
            ] coordinates {
                (No Col., 0.9550) (Driv. Area, 1.0000) (Direction, 0.9890)
                (Speed Lim., 0.9880) (Progress, 0.8850) (TTC, 0.9250)
                (Comfort, 0.9850) (Lane Foll., 0.9400) (Proximity, 0.9800)
            };
            \legend{DiffusionPlanner, RAPiD}

        % ---------------------------------------------------------
        % Plot (b): Test14
        % ---------------------------------------------------------
        \nextgroupplot[
            title={(b) Test14}, 
            title style={font=\scriptsize, at={(0.5,0.9)}},
            xticklabels=\empty
        ]
            \addplot[fill=customRed, draw=none, bar width=4pt, bar shift=-2.5pt] coordinates {
                (No Col., 0.9511) (Driv. Area, 1.0000) (Direction, 0.9812)
                (Speed Lim., 0.9727) (Progress, 0.9408) (TTC, 0.9249)
                (Comfort, 0.9823) (Lane Foll., 0.9258) (Proximity, 0.9753)
            };
            \addplot[fill=customGreen, draw=none, bar width=4pt, bar shift=2.5pt] coordinates {
                (No Col., 0.9595) (Driv. Area, 1.0000) (Direction, 0.9880)
                (Speed Lim., 0.9890) (Progress, 0.8920) (TTC, 0.9350)
                (Comfort, 0.9890) (Lane Foll., 0.9450) (Proximity, 0.9820)
            };

        % ---------------------------------------------------------
        % Plot (c): Test14-Hard
        % ---------------------------------------------------------
        \nextgroupplot[
            title={(c) Test14-Hard}, 
            title style={font=\scriptsize, at={(0.5,0.9)}},
        ]
            \addplot[fill=customRed, draw=none, bar width=4pt, bar shift=-2.5pt] coordinates {
                (No Col., 0.9199) (Driv. Area, 1.0000) (Direction, 0.9623)
                (Speed Lim., 0.9712) (Progress, 0.8977) (TTC, 0.8849)
                (Comfort, 0.9758) (Lane Foll., 0.9206) (Proximity, 0.9543)
            };
            \addplot[fill=customGreen, draw=none, bar width=4pt, bar shift=2.5pt] coordinates {
                (No Col., 0.9310) (Driv. Area, 1.0000) (Direction, 0.9690)
                (Speed Lim., 0.9820) (Progress, 0.8850) (TTC, 0.8950)
                (Comfort, 0.9800) (Lane Foll., 0.9250) (Proximity, 0.9610)
            };
            
        \end{groupplot}

    \end{tikzpicture}
    \caption{Breakdown of PDM scorer sub-metrics across (a) Val14, (b) Test14, and (c) Test14-Hard splits. RAPiD (Green) consistently outperforms DiffusionPlanner (Red) in critical safety metrics (\textit{No Collision}, \textit{TTC}) while trading \textit{Progress} for safer maneuvers. \textbf{Metric Definitions:} \textbf{No Col.}: No Collision, \textbf{Driv. Area}: Drivable Area, \textbf{Speed Lim.}: Speed Limit, \textbf{TTC}: Time-to-Collision, \textbf{Lane Foll.}: Lane Following.}
    \label{fig:all_splits_comparison}
\end{figure}
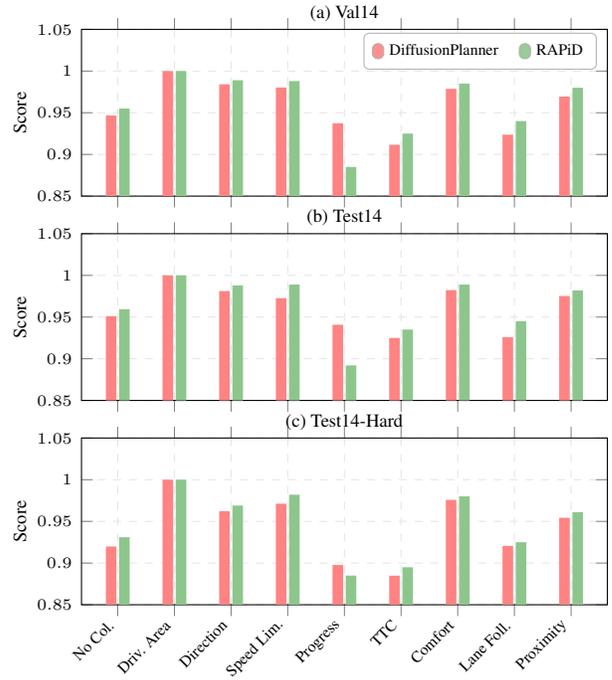

\noindent\textbf{Deployment Ablations.}
Table~\ref{tab:selection_ablation} isolates the contribution of
critic-based selection on nuPlan Val14-R. Randomly selecting one of
$K{=}32$ candidates performs similarly to generating a single candidate,
as expected because both return an unranked sample from the same
distribution. In contrast, critic-based selection improves CLS from
79.15 to 82.13 and improves progress, collision avoidance, and TTC,
showing that the gain arises from value-based ranking rather than
candidate generation alone.

Table~\ref{tab:nfe_ablation} evaluates the student sampling depth.
The configured two-step sampler achieves the best CLS and safety
metrics. One-step inference provides insufficient refinement, whereas
additional evaluations do not improve performance because the student
was distilled for the two-step schedule, while substantially increasing
decoder latency. We therefore use NFE$=2$ and $K{=}32$ for deployment.

\noindent \textbf{interPlan Benchmark Analysis.} We evaluate our approach against representative baselines, including learning-based planners (UrbanDriver~\cite{scheel2022urban}, GC-PGP~\cite{hallgarten2023prediction}, GameFormer~\cite{huang2023gameformer}, PDM-Open~\cite{dauner2023parting}, DTPP~\cite{huang2024dtpp}, DiffusionPlanner~\cite{zheng2025diffusion}) and rule-based planners (IDM~\cite{treiber2000congested}, IDM+MOBIL~\cite{kesting2007general}, 
PDM-Closed~\cite{dauner2023parting}).
The interPlan scorer extends standard nuPlan metrics (compliance, safety, and comfort) with scenario-specific requirements. It measures lane change completion rates relative to goal requirements and applies multiplicative progress penalties for failures to pass obstacles like parked cars or construction zones. Scenarios receive zero scores for collisions, drivable area violations, extended stationarity, or getting stuck behind obstacles. Direction compliance penalties are disabled for overtaking and accident scenarios where oncoming lane usage is necessary.

Table~\ref{tab:interPlan_scores} presents results on the interPlan benchmark, which targets scenarios that are under-emphasized in standard nuPlan evaluation: achieving high scores through simple lane-following with minimal interaction. interPlan augments scenario goals to require multiple lane changes and initializes varying traffic densities by spawning agents around the ego vehicle at maximum spacings of 100m (low traffic density, LTD), 50m (medium traffic density, MTD), and 33m (high traffic density, HTD).
The benchmark comprises 80 scenarios across eight categories testing planner generalization to long-tail situations. Lateral maneuvering is evaluated through construction zones (Const., navigating around traffic cones), accident sites (Acc., maneuvering around crashed vehicles), jaywalking pedestrians (Jayw., handling unexpected crossings at bus stops), nudging (carefully passing parked vehicles within lane), and overtaking (Overt., safe passage through oncoming lanes). Please see Section S7-B of \textit{Supplementary Material} for details. Longitudinal interaction is tested through lane changes under LTD, MTD, and HTD conditions with varying traffic environments (conservative, assertive, and mixed). Each category contains 10 instances. Compared with standard nuPlan evaluation, where conservative lane-following can achieve strong scores, interPlan explicitly requires interactive maneuvering, proactive gap selection, and robust generalization to uncommon situations.
Our approach achieves the highest interPlan score among learning-based methods with an aggregate score of 30, outperforming PDM-Open and DiffusionPlanner. For lateral maneuvers, our method scores 61 in nudging scenarios (versus DiffusionPlanner's 60) and 26 in overtaking, \ie, the highest among both learning-based and rule-based methods. For longitudinal interactions, performance improvements increase with traffic density, \ie, 42 versus 40 in LTD, 27 versus 21 in MTD, and 49 versus 48 in HTD. These gains in denser conditions demonstrate our method's ability to navigate between vehicles and complete required lane changes without freezing or causing safety violations. In MTD and HTD scenarios with reduced agent spacing (50m and 33m), our method shows improved gap selection and timing compared to DiffusionPlanner. While rule-based methods achieve highest performance on nuPlan, they fail to meet interPlan requirements in several scenarios. Our learning-based model demonstrates the adaptability required for interactive scenarios while maintaining competitive performance.

\noindent \textbf{PDM-Based Safety and Comfort Analysis.}
Fig.~\ref{fig:all_splits_comparison} provides a metric-level breakdown of PDM scores in the reactive environment across Val14, Test14, and Test14-hard splits. Both \mymethod~and DiffusionPlanner are evaluated in the same reactive closed-loop setting, but this analysis reports the PDM-style safety and comfort metrics rather than the standard nuPlan score. The results help explain the reactive-score trade-off observed in Table~\ref{tab:nuPlan_benchmark}, \ie, \mymethod~improves several safety- and comfort-oriented metrics while reducing progress. \newline
\textbf{Safety Metrics.}
\mymethod~achieves higher collision-avoidance scores across all three splits. The no-collision score increases from 0.9467 to 0.9550 on Val14, from 0.9511 to 0.9595 on Test14, and from 0.9199 to 0.9310 on Test14-hard. Both methods maintain perfect drivable-area compliance. \mymethod~also maintains high driving-direction compliance, with scores of 0.9890, 0.9880, and 0.9690 across Val14, Test14, and Test14-hard. Speed-limit compliance improves across all three splits, increasing from 0.9803 to 0.9880 on Val14, from 0.9727 to 0.9890 on Test14, and from 0.9712 to 0.9820 on Test14-hard. Time-to-collision (TTC) also improves from 0.9116 to 0.9250 on Val14, from 0.9249 to 0.9350 on Test14, and from 0.8849 to 0.8950 on Test14-hard. These improvements indicate that the PDM-trained critic tends to select trajectories with larger safety margins. \newline
\textbf{Comfort, Proximity, and Lane-Following Metrics.}
\mymethod~also improves comfort, proximity, and lane-following scores. On Val14, comfort increases from 0.9787 to 0.9850; on Test14, from 0.9823 to 0.9890; and on Test14-hard, from 0.9758 to 0.9800. Proximity scores similarly improve from 0.9694 to 0.9800 on Val14, from 0.9753 to 0.9820 on Test14, and from 0.9543 to 0.9610 on Test14-hard. Lane-following improves from 0.9237 to 0.9400 on Val14, from 0.9258 to 0.9450 on Test14, and from 0.9206 to 0.9250 on Test14-hard. These trends are consistent with the modified PDM-style reward used for critic training, where the comfort weight is increased from 2 to 5 and a proximity-to-leading-agent term is introduced with weight 5. \newline
\textbf{Progress Metrics.}
The main trade-off appears in progress. Progress decreases from 0.9374 to 0.8850 on Val14, from 0.9408 to 0.8920 on Test14, and from 0.8977 to 0.8850 on Test14-hard. This reduction is consistent with the reward design, where the progress weight is reduced from 5 to 2 while comfort, proximity, TTC, and collision avoidance receive stronger emphasis. As a result, \mymethod~tends to prefer more conservative trajectories with improved safety and comfort margins, which can lower the standard reactive nuPlan score when progress is weighted more heavily.

\section{Conclusion}

    This work addresses the computational inefficiency and stochastic behavior of diffusion-based trajectory planning through efficient multimodal policy distillation. Rather than extracting a single deterministic action, which collapses the teacher's multimodality and limits reactive closed-loop performance, we distill a pretrained DiffusionPlanner model into a few-step consistency student and deploy it with critic best-of-$K$ selection, preserving the teacher's behavioral diversity while training with PDM-based safety scoring. By anchoring the student to the prior's probability-flow ODE (Ordinary Differential Equation), generated trajectories remain within the support of valid driving behaviors, and the 2-step sampler reads out where the denoiser is accurate, yielding safe and comfortable trajectories at real-time inference. Evaluation on nuPlan and interPlan benchmarks demonstrates that our method matches the diffusion teacher on the non-reactive nuPlan splits and generalizes strongly among learning-based methods on interPlan, while delivering a $5.5\times$ speedup in complete-pipeline inference latency (18.41 ms versus 100.91 ms) compared to the diffusion baseline. On the reactive splits our method remains competitive with the teacher, with a small difference in score. Our work provides a practical pathway for deploying efficient, safety-focused planning systems in real-world autonomous driving by leveraging pretrained diffusion models without requiring costly training from scratch, bridging the gap between diffusion model expressiveness and the latency requirements of production autonomous vehicles.

% \bibliographystyle{IEEEtran}
% \bibliography{ref}

% --- Start Supplementary ---
\twocolumn[{
\centering
{\LARGE \bfseries Supplementary Material\par}
\vspace{6.8em}
}]

% Resetting section numbering to S1, S2 as per your requirement
\renewcommand{\thesection}{S\arabic{section}}
\setcounter{section}{0}

% Resetting figure numbering to S1, S2
\renewcommand{\thefigure}{S\arabic{figure}}
\setcounter{figure}{0}
%\section*{Supplementary Material}
% \documentclass[lettersize,journal]{IEEEtran}
% \usepackage{amsmath,amsfonts}
% \usepackage{algorithmic}
% \usepackage{algorithm}
% \usepackage{array}
% \usepackage[caption=false,font=normalsize,labelfont=sf,textfont=sf]{subfig}
% \usepackage{textcomp}
% \usepackage{stfloats}
% \usepackage{url}
% \usepackage{verbatim}
% \usepackage{graphicx}
% \usepackage{cite}
% \usepackage{times}
% \usepackage{soul}
% \usepackage{url}
% \usepackage[hidelinks]{hyperref}
% \usepackage[utf8]{inputenc}
% \usepackage[small]{caption}
% \usepackage{graphicx}
% \usepackage{amsmath}
% \usepackage{amsthm}
% \usepackage{xspace}
% \usepackage{xcolor}
% \usepackage{booktabs}
% \usepackage{algorithm}
% \usepackage{algorithmic}
% \usepackage[switch]{lineno}
% \usepackage{multirow}
\makeatletter
\DeclareRobustCommand\onedot{\futurelet\@let@token\@onedot}
\def\@onedot{\ifx\@let@token.\else.\null\fi\xspace}
\def\eg{\emph{e.g}\onedot} \def\Eg{\emph{E.g}\onedot}
\def\ie{\emph{i.e}\onedot} \def\Ie{\emph{I.e}\onedot}
\makeatother
\pgfplotsset{compat=1.14}
\usepgfplotslibrary{groupplots}
\usetikzlibrary{patterns}

\hyphenation{op-tical net-works semi-conduc-tor IEEE-Xplore}
% updated with editorial comments 8/9/2021
% \newcommand{\mymethod}{RAPiD}
\renewcommand{\thesection}{S\arabic{section}}
% \begin{document}

\title{RAPiD: Reward-Guided Consistency Distillation of Diffusion Planners for Real-Time Autonomous Driving}

\maketitle

\section{Basics of Predictive Driver Model (PDM)}
\begin{figure}[ht]
    \centering
    \includegraphics[width=\linewidth]{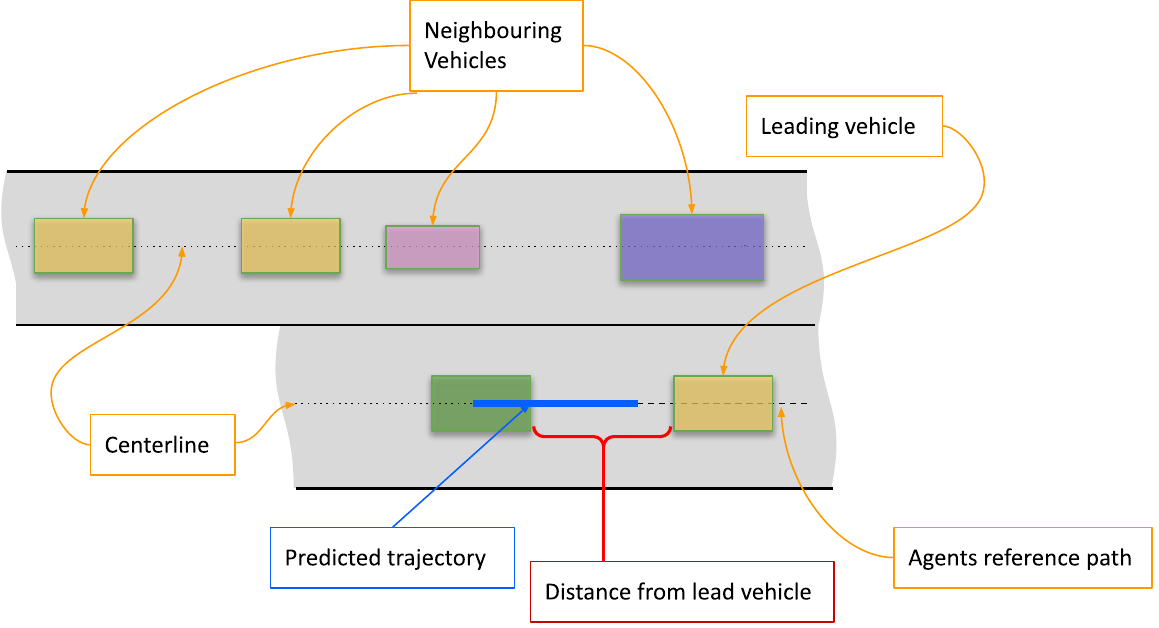}
    \caption{Supplementary overview of PDM metrics.}
    \label{fig:PDM_overview}
\end{figure}

\begin{figure}[ht]
    \centering
    \includegraphics[width=\linewidth]{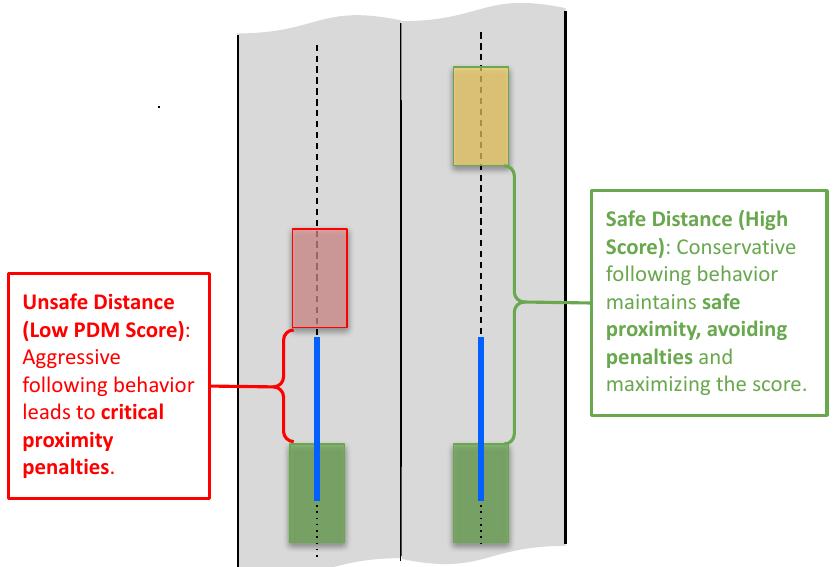}
    \caption{Visualisation of PDM safety distance metrics.}
    \label{fig:safe_dist}
\end{figure}

\begin{figure}[ht]
    \centering
    \includegraphics[width=\linewidth]{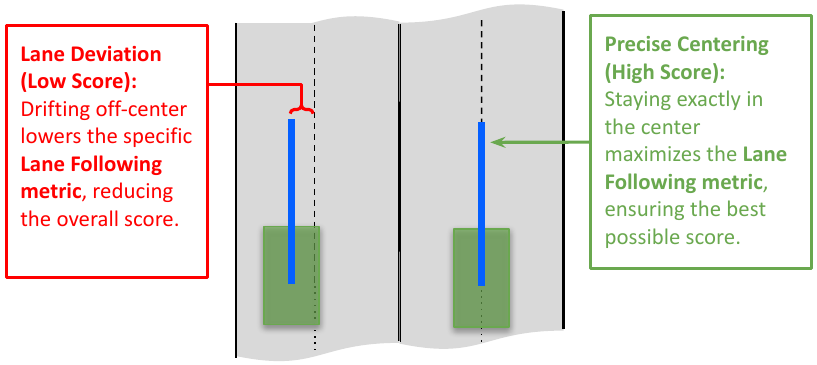}
    \caption{Analysis of Centerline tracking performance.}
    \label{fig:lane_center}
\end{figure}

\begin{figure}[ht]
    \centering
    \includegraphics[width=\linewidth]{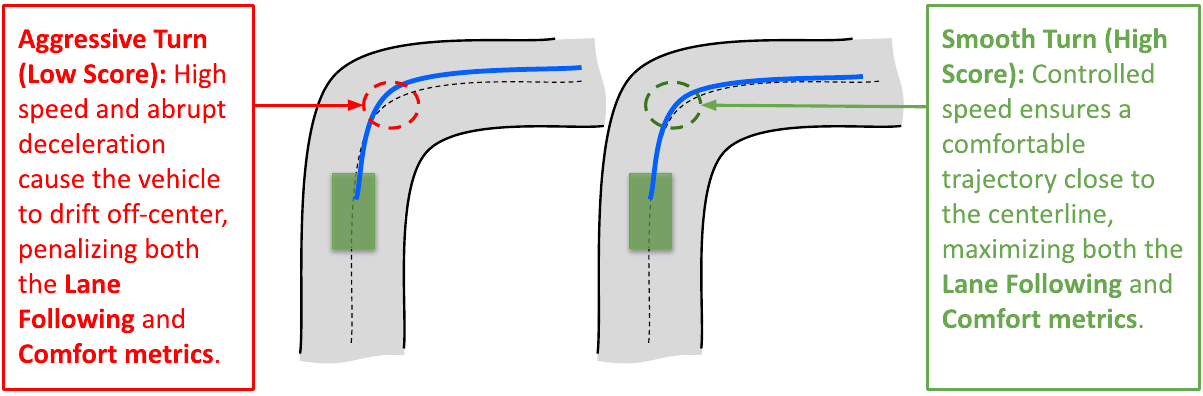}
    \caption{Qualitative evaluation of turning behavior.}
    \label{fig:turn}
\end{figure}

This Section compliments Section III-D of the main paper. PDM~\cite{dauner2023parting,diffusionES} evaluates trajectory safety and quality through a composite of weighted metrics. While the full PDM Scorer includes continuous metrics such as speed limit compliance and comfort, Figure~\ref{fig:PDM_overview} provides a supplementary overview of the primary spatial components used for trajectory selection. This includes monitoring the distance from the leading vehicle, divergence from the lane centerline, and adherence to the agent's reference path. A critical safety component is the maintenance of a Safe Distance from the leading vehicle, as shown in Figure~\ref{fig:safe_dist}. In the PDM framework, this is governed by high multiplier weights for Proximity and TTC (Time-to-Collision) within bounds, which penalize aggressive following behavior to minimize collision risk. As illustrated in Figure~\ref{fig:lane_center}, Centerline Tracking is essential for maintaining Lane Following and Drivable Area Compliance. Precise centering ensures adequate clearance from neighboring vehicles, avoiding the penalties associated with drifting off-center. Finally, Figure~\ref{fig:turn} highlights Turning Behaviour, where Comfort metrics (weighted at 5 in PDM) penalize the aggressive lateral and longitudinal accelerations caused by late deceleration or overspeeding during maneuvers.

% Plot 1: Supplementary Overview
% Plot 2: PDM Visualisation
% Plot 3: Centerline Tracking
% Plot 4: Turning Behavior

\section{Method Pseudocode}
This Section compliments Section III-E of the main paper.
The complete framework is detailed in Algorithm~\ref{alg:cd_training} (the three-stage training pipeline) and Algorithm~\ref{alg:cd_deploy} (deployment). Stage 1 constructs the offline replay buffer from a balanced mixture of PDM-scored ground-truth log-replay and DiffusionPlanner rollout transitions, Stage 2 trains the Implicit Q-Learning (IQL) critic, and Stage 3 distills a few-step consistency student from the frozen DiffusionPlanner~\cite{zheng2025diffusion} prior. At deployment, the student draws $K$ candidate trajectories with a 2-step sampler and acts with the critic-selected best.

\begin{algorithm}
\caption{Consistency Distillation Training Pipeline}
\label{alg:cd_training}
    \begin{algorithmic}[1]
    \REQUIRE Frozen DiffusionPlanner encoder $\mathcal{E}$ and denoiser $\epsilon_\psi$ (behavior prior)
    \REQUIRE PDM Scorer $\mathcal{S}_{\text{PDM}}$; balanced raw replay dataset
    $\mathcal{D}_{\text{raw}}=\{(o_k,a_k,o_{k+1})\}$ collected from ground-truth log replay and DiffusionPlanner rollouts
    \REQUIRE Hyperparameters: $\tau$ (expectile), $\gamma$; noise schedule
    $(\alpha_t,\sigma_t)$; grid $\{t_n\}_{n=0}^{N}$; weights
    $w_q,\lambda,w_{\text{dsm}}$; EMA rate $\mu$; anchor range
    $t_{\max}^{\text{dsm}}$; max time $T$
    
    \STATE \textbf{Stage 1: Offline Replay Buffer Creation}
    \STATE Initialize replay buffer $\mathcal{D}\leftarrow\emptyset$
    \FOR{each $(o_k,a_k,o_{k+1})\in\mathcal{D}_{\text{raw}}$}
        \STATE Encode states:
        $s_k\leftarrow\mathcal{E}(o_k)$;
        \quad $s_{k+1}\leftarrow\mathcal{E}(o_{k+1})$
        \COMMENT{frozen encoder}
        \STATE Compute reward:
        $r_k\leftarrow\mathcal{S}_{\text{PDM}}(o_k,a_k)$
        \COMMENT{safety \& comfort}
        \STATE Store:
        $\mathcal{D}\leftarrow\mathcal{D}\cup
        \{(s_k,s_{k+1},a_k,r_k)\}$
    \ENDFOR

    \STATE \textbf{Stage 2: Critic Training (IQL)}
    \STATE Initialize value $V_\zeta$ and twin critics $Q_{\phi,1}, Q_{\phi,2}$; target $Q_{\phi^-}$
    \WHILE{not converged}
        \STATE Sample batch $B \sim \mathcal{D}$
        \STATE \textit{// Value update (expectile regression)}
        \STATE $\mathcal{L}_V(\zeta) \leftarrow \mathbb{E}_B\!\left[ L_2^\tau\!\left( Q_{\phi^-}(s,a) - V_\zeta(s) \right) \right]$
        \STATE $\zeta \leftarrow \zeta - \alpha_V \nabla_\zeta \mathcal{L}_V$
        \STATE \textit{// Q update (TD error)}
        \STATE $y \leftarrow r + \gamma\,V_\zeta(s')$
        \STATE $\mathcal{L}_Q(\phi) \leftarrow \mathbb{E}_B\!\left[ \textstyle\sum_{i=1,2} \| y - Q_{\phi,i}(s,a) \|_2^2 \right]$
        \STATE $\phi \leftarrow \phi - \alpha_Q \nabla_\phi \mathcal{L}_Q$; \quad update target $Q_{\phi^-}$
    \ENDWHILE

    \STATE \textbf{Stage 3: Consistency Distillation}
    \STATE Initialize student $f_\theta$ (EDM preconditioning); EMA target $\theta^- \leftarrow \theta$
    \WHILE{not converged}
        \STATE Sample batch $(s, a_0) \sim \mathcal{D}$
        \STATE \textit{// Critic guidance on one-step readout}
        \STATE $z \sim \mathcal{N}(0,I)$; \quad $a_{\text{gen}} \leftarrow \mathrm{clip}\!\left( f_\theta(\sigma_T z, T, s),\, \pm 1 \right)$
        \STATE $q \leftarrow \tfrac{1}{2}\!\left( Q_{\phi^-,1} + Q_{\phi^-,2} \right)(s, a_{\text{gen}})$
        \STATE $\mathcal{L}_Q \leftarrow -\,\mathbb{E}\!\left[\, q \,/\, \overline{|q|}\, \right]$ \COMMENT{value normalized by $\overline{|q|}$ for scale-invariance}
        \STATE \textit{// Consistency loss (adjacent grid pair $t_{n+1} > t_n$)}
        \STATE $n \sim \mathcal{U}\{0,\dots,N-1\}$; \quad $\epsilon \sim \mathcal{N}(0,I)$
        \STATE $x_{n+1} \leftarrow \alpha_{n+1} a_0 + \sigma_{n+1}\epsilon$
        \STATE $\hat\epsilon \leftarrow \epsilon_\psi(x_{n+1}, t_{n+1}, s)$;\; $\hat a_0 \leftarrow (x_{n+1} - \sigma_{n+1}\hat\epsilon)/\alpha_{n+1}$
        \STATE $\hat x_n \leftarrow \alpha_n \hat a_0 + \sigma_n \hat\epsilon$ \COMMENT{teacher DDIM step, stop-grad}
        \STATE $\mathcal{L}_{\text{CD}} \leftarrow w(t_n)\, d\!\left( f_\theta(x_{n+1}, t_{n+1}, s),\, f_{\theta^-}(\hat x_n, t_n, s) \right)$
        \STATE \textit{// Low-noise data anchor}
        \STATE $t_d \sim \mathcal{U}(t_\epsilon, t_{\max}^{\text{dsm}})$;\; $x_d \leftarrow \alpha_d a_0 + \sigma_d \epsilon_d$
        \STATE $\mathcal{L}_{\text{DSM}} \leftarrow d\!\left( f_\theta(x_d, t_d, s),\, a_0 \right)$
        \STATE \textit{// Combined update and EMA}
        \STATE $\mathcal{L} \leftarrow w_q \mathcal{L}_Q + \lambda \mathcal{L}_{\text{CD}} + w_{\text{dsm}} \mathcal{L}_{\text{DSM}}$
        \STATE $\theta \leftarrow \theta - \eta \nabla_\theta \mathcal{L}$
        \STATE $\theta^- \leftarrow \mu \theta^- + (1-\mu)\theta$
    \ENDWHILE
    \RETURN Trained student $f_\theta$
    \end{algorithmic}
\end{algorithm}

\begin{algorithm}
\caption{Deployment: Best-of-$K$ with 2-Step Sampler}
\label{alg:cd_deploy}
    \begin{algorithmic}[1]
    \REQUIRE Student $f_\theta$; twin critics $Q_{\phi^-,1}, Q_{\phi^-,2}$; encoder $\mathcal{E}$
    \REQUIRE Candidates $K$; generation steps (NFE) $=2$; floor time $t_{\text{floor}}$
    \STATE Encode state: $s \leftarrow \mathcal{E}(s_{\text{raw}})$
    \FOR{$k = 1$ to $K$}
        \STATE $z \sim \mathcal{N}(0,I)$;\; $a \leftarrow f_\theta(\sigma_T z, T, s)$
        \FOR{$t$ in interior nodes of $\text{linspace}(T, t_{\text{floor}}, \text{steps}{+}1)$}
            \STATE $z' \sim \mathcal{N}(0,I)$;\; $x \leftarrow \alpha_t a + \sigma_t z'$;\; $a \leftarrow f_\theta(x, t, s)$
        \ENDFOR
        \STATE $a_k \leftarrow \mathrm{clip}(a, \pm 1)$
    \ENDFOR
    \STATE \textbf{Select} $a^\star \leftarrow \arg\max_{a_k} \tfrac{1}{2}\!\left( Q_{\phi^-,1} + Q_{\phi^-,2} \right)(s, a_k)$
    \RETURN $a^\star$
    \end{algorithmic}
\end{algorithm}

\section{Consistency Distillation: Full Formulation}
This Section complements Section III-E of the main paper and provides the complete mathematical specification of the deployed discrete consistency student, verified against our implementation.

\subsection{Teacher Noise Process (VP-SDE)}
The frozen DiffusionPlanner teacher is trained under a variance-preserving (VP) diffusion process. For a clean trajectory $a_0$ and continuous time $t\in[t_\epsilon, T]$ ($T=0.98$), the forward marginal is $x_t = \alpha_t a_0 + \sigma_t\epsilon$, $\epsilon\sim\mathcal{N}(0,I)$, with
\begin{equation}
\log\alpha_t = -\tfrac{1}{4}t^2(\beta_1-\beta_0) - \tfrac{1}{2}t\,\beta_0, \quad
\alpha_t = e^{\log\alpha_t}, \quad
\sigma_t = \sqrt{1-\alpha_t^2},
\end{equation}
where $\beta_0=0.1$, $\beta_1=20$, so that $\alpha_t^2+\sigma_t^2=1$. The teacher $\epsilon_\psi(x_t,t,s)$ predicts the noise conditioned on the latent state $s=\mathcal{E}(s_{\text{raw}})$ and is held frozen throughout distillation.

\subsection{Student Parameterization (EDM Preconditioning)}
The consistency student $f_\theta(x_t,t,s)$ maps a noised trajectory back to a clean one. We adopt EDM-style preconditioning~\cite{karras2022elucidating} adapted to the VP process. Writing the variance-exploding-equivalent variable $\tilde{x} = x_t/\alpha_t = a_0 + \tilde\sigma\,\epsilon$ with $\tilde\sigma = \sigma_t/\alpha_t$ and data scale $\sigma_d=0.5$,
\begin{equation}
f_\theta(x_t,t,s) = c_{\text{skip}}\,\tilde{x} + c_{\text{out}}\,F_\theta(c_{\text{in}}\tilde{x},\,t,\,s),
\end{equation}
\begin{equation}
c_{\text{skip}} = \frac{\sigma_d^2}{\tilde\sigma^2+\sigma_d^2}, \quad
c_{\text{out}} = \frac{\tilde\sigma\,\sigma_d}{\sqrt{\tilde\sigma^2+\sigma_d^2}}, \quad
c_{\text{in}} = \frac{1}{\sqrt{\tilde\sigma^2+\sigma_d^2}},
\end{equation}
where $F_\theta$ is the trainable network. As $t\to 0$, $\tilde\sigma\to 0$ gives $c_{\text{skip}}\to 1$ and $c_{\text{out}}\to 0$, enforcing the consistency boundary condition $f_\theta(x,0)=x$. This boundary is essential: it prevents the consistency loss from collapsing to a trivial trajectory-constant solution and keeps the network input unit-scaled at every noise level.

\subsection{Teacher Probability-Flow ODE Step}
The consistency target is anchored to one deterministic step of the teacher's probability-flow ODE, implemented as a DDIM step~\cite{song2020denoising} from a noisier node $t_{n+1}$ to an adjacent less-noisy node $t_n$. Given $x_{n+1}$, this step is computed under stop-gradient:
\begin{equation}
\hat\epsilon = \epsilon_\psi(x_{n+1}, t_{n+1}, s), \;\;
\hat a_0 = \frac{x_{n+1}-\sigma_{n+1}\hat\epsilon}{\alpha_{n+1}}, \;\;
\hat x_n = \alpha_n\hat a_0 + \sigma_n\hat\epsilon.
\end{equation}
Gradients are never propagated through this teacher step; it serves only to produce the regression target $\hat x_n$.

\subsection{Distillation Time Grid}
Distillation is performed on an ordered grid $t_\epsilon = t_0 < t_1 < \dots < t_N = T$ of $N{+}1$ nodes ($N=20$). The default grid is uniform in $t$; an optional grid uniform in log-SNR, $\mathrm{logSNR}(t)=2\log\alpha_t-2\log\sigma_t$, equalizes the per-step teacher-ODE accuracy across nodes. At each step, an index $n\sim\mathcal{U}\{0,\dots,N-1\}$ selects the adjacent pair $(t_{n+1},t_n)$.

\subsection{Training Objective}
The student minimizes a three-term objective,
\begin{equation}
\mathcal{L}(\theta) = w_q\,\mathcal{L}_Q + \lambda\,\mathcal{L}_{\text{CD}} + w_{\text{dsm}}\,\mathcal{L}_{\text{DSM}},
\end{equation}
combining critic guidance, prior-ODE self-consistency, and a low-noise data anchor.

\noindent\textbf{Consistency term.} For the sampled adjacent pair, with $x_{n+1}=\alpha_{n+1}a_0+\sigma_{n+1}\epsilon$ and the teacher-evolved target $\hat x_n$,
\begin{equation}
\mathcal{L}_{\text{CD}} = \mathbb{E}\big[\, w(t_n)\, d\big(f_\theta(x_{n+1},t_{n+1},s),\; f_{\theta^-}(\hat x_n,t_n,s)\big) \big],
\end{equation}
where $\theta^-$ is an exponential-moving-average (EMA) target, updated as $\theta^-\leftarrow\mu\theta^-+(1-\mu)\theta$ with $\mu=0.999$ and held under stop-gradient. The target network is evaluated in inference mode so that dropout does not perturb the regression target. The per-step weight defaults to $w(t_n)=1$ (an optional SNR weighting $\alpha_n^2/\sigma_n^2$ is also supported). This term keeps the student on the prior's probability-flow ODE, replacing iterative sampling while preserving the teacher's support over valid driving behaviors.

\noindent\textbf{Data anchor.} To stabilize the high-noise consistency bootstrap, the student's clean prediction is regressed directly to ground-truth trajectories at low noise $t_d\sim\mathcal{U}(t_\epsilon, t_{\max}^{\text{dsm}})$, with $x_d=\alpha_d a_0+\sigma_d\epsilon_d$:
\begin{equation}
\mathcal{L}_{\text{DSM}} = \mathbb{E}\,d\big(f_\theta(x_d,t_d,s),\; a_0\big).
\end{equation}
Restricting this anchor to low noise provides a strong on-manifold signal without inducing the mode-averaging that regressing to the mean at high noise would cause.

\noindent\textbf{Critic guidance.} The reward term is computed on the one-step readout from pure noise, $a_{\text{gen}}=\mathrm{clip}(f_\theta(\sigma_T z, T, s),\pm 1)$ with $z\sim\mathcal{N}(0,I)$, scored by the mean of the twin target critics:
\begin{equation}
\mathcal{L}_Q = -\,\mathbb{E}\!\left[\frac{\tfrac{1}{2}\big(Q_{\phi^-\!,1}+Q_{\phi^-\!,2}\big)(s,a_{\text{gen}})}{\overline{|Q|}}\right].
\end{equation}
The clip is load-bearing: its zero sub-gradient outside $[-1,1]$ prevents value ascent from driving $a_{\text{gen}}$ off-manifold into the critic's out-of-distribution extrapolation regime. Normalizing by the running mean magnitude $\overline{|Q|}$ makes the reward weight $w_q$ comparable across environments. Following reward-aware consistency distillation~\cite{duan2025accelerating}, $w_q$ is kept small so that distillation dominates and the critic only steers among in-support modes.

\noindent\textbf{Distance.} The default distance $d$ is a per-dimension Huber loss ($\delta{=}1$) summed over action dimensions; squared-error and an iCT-style pseudo-Huber $\sqrt{\lVert\cdot\rVert^2+c^2}-c$~\cite{song2024improved} are supported as alternatives.

\subsection{Deployment}
At inference the student is sampled with $\text{NFE}=2$: a one-step readout at $t=T$ followed by a renoise to a lower interior node and a second readout, where the denoiser is more accurate (Algorithm~\ref{alg:cd_deploy}). Because a single deterministic readout averages opposing modes of the multimodal posterior, we draw $K$ candidates and act with the critic-selected best, $a^\star = \arg\max_{a_k}\tfrac{1}{2}(Q_{\phi^-\!,1}+Q_{\phi^-\!,2})(s,a_k)$. As established in the main paper, performance saturates at $K=32$, the deployment setting.

\section{D4RL Benchmark Results}
This Section complements Section IV of the main paper. We evaluate the deployed discrete consistency student on D4RL~\cite{fu2020d4rl} to validate the distillation mechanism in a controlled setting. AntMaze is the closest D4RL proxy for closed-loop driving: multimodal, long-horizon, and sparse-reward, requiring the policy to commit to one of several valid routes rather than average across them. Locomotion (Gym-MuJoCo) is the complementary unimodal-precision regime. All results use best-of-$K$ deployment ($K{=}32$, 2-step sampler).

\begin{table*}[t]
\caption{Evaluation numbers of D4RL benchmarks (normalized as suggested by ~\cite{fu2020d4rl}). We report mean $\pm$ standard deviation where available (multi-seed) and single-seed otherwise for our method. Numbers within 5\% of the maximum in every individual task are highlighted.}
\label{tbl:rl_results}
\centering
\small
\resizebox{1.0\textwidth}{!}{%
\begin{tabular}{llccccccccccc}
\toprule
\multicolumn{1}{c}{\bf Dataset} & \multicolumn{1}{c}{\bf Environment} & \multicolumn{1}{c}{\bf BEAR} & \multicolumn{1}{c}{\bf TD3+BC} & \multicolumn{1}{c}{\bf IQL} & \multicolumn{1}{c}{\bf SfBC} & \multicolumn{1}{c}{\bf Diffuser} & \multicolumn{1}{c}{\bf Diffusion-QL}& \multicolumn{1}{c}{\bf QGPO} & \multicolumn{1}{c}{\bf IDQL} & \multicolumn{1}{c}{\bf SRPO} & \multicolumn{1}{c}{\bf Ours} \\
\midrule
Medium-Expert & HalfCheetah    &  $53.4$ &  $90.7$ &  $86.7$ & $\bf{92.6}$ & $79.8$ &  $\bf{96.8}$ & $\bf{93.5}$ & $\bf{95.9}$ & $\bf{92.2\pm3.0}$ & $42.5$ \\
Medium-Expert & Hopper         &  $96.3$ &  $98.0$ &  $91.5$ & $\bf{108.6}$ & $\bf{107.2}$ & $\bf{111.1}$ & $\bf{108.0}$ & $\bf{108.6}$ & $100.1\pm13.9$ & $65.1$ \\
Medium-Expert & Walker2d       & $40.1$  &  $\bf{110.1}$ & $\bf{109.6}$ & $\bf{109.8}$ & $\bf{108.4}$ & $\bf{110.1}$ & $\bf{110.7}$ & $\bf{112.7}$ & $\bf{114.0\pm2.1}$ & $\bf{111.2}$ \\
\midrule
Medium        & HalfCheetah    &  $41.7$ &  $48.3$ &  $47.4$ & $45.9$ & $44.2$ &  $51.1$ & $54.1$ & $51.0$ & $\bf{60.4\pm0.8}$ & $54.0$ \\
Medium        & Hopper         &  $52.1$ &  $59.3$ &  $66.3$ & $57.1$ & $58.5$ & $90.5$ & $\bf{98.0}$ & $65.4$ & $\bf{95.5\pm2.0}$ & $57.3$ \\
Medium        & Walker2d       &  $59.1$ &  $\bf{83.7}$ &  $78.3$ & $77.9$ & $79.7$ & $\bf{87.0}$ & $\bf{86.0}$ & $82.5$ & $\bf{84.4\pm4.4}$ & $82.9$ \\
\midrule
Medium-Replay & HalfCheetah    &  $38.6$ &  $44.6$ &  $44.2$ &   $37.1$ & $42.2$ & $47.8$ & $47.6$ & $45.9$ & $\bf{51.4\pm3.4}$ & $44.8$ \\
Medium-Replay & Hopper         &  $33.7$ &  $60.9$ &  $94.7$ &   $86.2$ & $\bf{101.3}$ & $\bf{100.7}$ & $\bf{96.9}$ & $92.1$ & $\bf{101.2\pm1.0}$ & $53.9$ \\
Medium-Replay & Walker2d       &  $19.2$ &  $81.8$ &  $73.9$ &   $65.1$ & $61.2$ &  $\bf{95.5}$ & $84.4$ & $85.1$ & $84.6\pm7.1$ & $81.2$ \\
\midrule
\multicolumn{2}{c}{\bf Average (Locomotion)} &  $51.9$ & $75.3$ & $76.9$ & $75.6$ & $75.3$ & $\bf{88.0}$ & $\bf{86.6}$ & $82.1$ & $\bf{87.1}$ & $65.9$ \\
\midrule
Default       & AntMaze-umaze  &  $73.0$ & $78.6$ & $87.5$ & $92.0$ & - & $93.4$ & $\bf{96.4}$ & $\bf{94.0}$ & $\bf{97.1\pm2.7}$ & $\bf{97.6}$ \\
Diverse       & AntMaze-umaze  &  $61.0$ & $71.4$ & $62.2$ & $\bf{85.3}$ & - & $66.2$ & $74.4$ & $80.2$ & $\bf{82.1\pm10.8}$ & $\bf{94.8}$ \\
\midrule
Play          & AntMaze-medium &  $0.0$ & $10.6$ & $71.2$ & $\bf{81.3}$ & - & $76.6$ & $\bf{83.6}$ & $\bf{84.5}$ & $80.7\pm7.1$ & $76.0$ \\
Diverse       & AntMaze-medium &  $8.0$ & $3.0$ & $70.0$ & $\bf{82.0}$ & - & $78.6$ & $\bf{83.8}$ & $\bf{84.8}$ & $75.0\pm12.3$ & $\bf{83.7}$ \\
\midrule
Play          & AntMaze-large  &  $0.0$ & $0.2$ & $39.6$ & $59.3$ & - & $46.4$ & $\bf{66.6}$ & $\bf{63.5}$ & $53.6\pm12.5$ & $59.4$ \\
Diverse       & AntMaze-large  &  $0.0$ & $0.0$ & $47.5$ & $45.5$ & - & $56.6$ &  $\bf{64.8}$ & $\bf{67.9}$ & $53.6\pm6.3$ & $45.6$ \\
\midrule
\multicolumn{2}{c}{\bf Average (AntMaze)} &  $23.7$ & $27.3$ & $63.0$ & $74.2$ & - & $69.6$ & $\bf{78.3}$ & $\bf{79.1}$ & $73.6$ & $\bf{76.2}$ \\
\bottomrule
\end{tabular}
}
\end{table*}

\subsection{Discussion}
Three findings emerge. First, the multimodal regime is where the consistency student delivers. On AntMaze its aggregate score ($76.2$) exceeds SRPO~\cite{chen2024score} ($73.6$) and is competitive with the strongest sampling-based methods, IDQL ($79.1$) and QGPO ($78.3$), despite acting in only two function evaluations rather than a full diffusion rollout. The student is strongest on the diverse mazes, reaching $94.8$ on umaze-diverse and $83.7$ on medium-diverse, both above SRPO ($82.1$ and $75.0$). This validates the central claim that a few-step multimodal student recovers the decision quality a single deterministic action sacrifices, at near-deterministic inference cost.

Second, the gains are largest precisely where multiple valid routes must be distinguished (the diverse splits), and the student remains within range of the best methods on the play and large mazes. This is consistent with the mechanism: the benefit of a multimodal student over a deterministic policy grows with the number of competing modes the policy must represent.

Third, locomotion is the clear limitation. The student trails SRPO by a wide aggregate margin ($65.9$ vs $87.1$), concentrated almost entirely in the hopper and halfcheetah-expert tasks. These are unimodal-precision problems: the dataset contains one near-optimal behavior, and a deterministic score-regularized policy reproduces it more faithfully than a stochastic few-step sampler, whose injected renoise adds variance with no multimodal benefit to offset it. This is not a failure of distillation but a regime mismatch, and it directly motivates the continuous-time student of Section~\ref{sec:scm}, which is built for precisely this regime, together with the reward-guidance instability that confines it there.

\section{Ablations}
This Section complements Section IV of the main paper and isolates the design choices the deployed student depends on: critic-based selection, the 2-step sampler, and the distillation grid. All ablations are run on D4RL AntMaze, the multimodal proxy for closed-loop driving, with $K{=}32$ candidates unless stated otherwise.

\subsection{Selection Rule: The Critic Ranking Is the Gain}
We first ask whether the improvement comes from drawing multiple candidates or from ranking them with the critic. Table~\ref{tab:abl_selection} compares acting on a single sample, selecting a random candidate from $K$, and selecting the critic-argmax candidate. Drawing $K$ candidates without critic ranking is no better than a single sample (random-of-32 $\approx$ single across every maze); the entire gain comes from the critic selecting the highest-value mode, lifting medium-diverse from $7.8$ to $81.9$ and large-play from $1.8$ to $57.8$. This directly supports the upgrade from a deterministic single action to critic-guided multimodal selection.

\begin{table}[ht]
    \centering
    \caption{Selection-rule ablation on AntMaze (normalized score $\times 100$, 2-step, $K{=}32$). The critic ranking, not the sampling of $K$ candidates, delivers the gain.}
    \label{tab:abl_selection}
    \begin{tabular}{lccc}
        \toprule
        \textbf{Maze} & \textbf{single ($K{=}1$)} & \textbf{random-of-32} & \textbf{critic-of-32} \\
        \midrule
        medium-play    & $4.6$ & $7.8$ & $\mathbf{73.4}$ \\
        medium-diverse & $7.8$ & $6.5$ & $\mathbf{81.9}$ \\
        large-play     & $1.8$ & $1.0$ & $\mathbf{57.8}$ \\
        large-diverse  & $1.3$ & $1.8$ & $\mathbf{48.9}$ \\
        \bottomrule
    \end{tabular}
\end{table}

\subsection{Generation Steps: 2-Step Is the Sweet Spot}
We next vary the number of denoise--renoise sampling steps (NFE). Table~\ref{tab:abl_nfe} shows that a single step fails entirely on every maze (score $0.0$), 2-step is best or tied, and additional steps yield no further gain. The 1-step failure is an accuracy failure, not a diversity failure: a single readout at the maximum noise level averages opposing modes of the multimodal posterior into an invalid action. The second step renoises to a lower noise level and reads out where the denoiser is accurate, placing the action on a real mode. This justifies the fast 2-step deployment.

\begin{table}[ht]
    \centering
    \caption{Generation-step (NFE) ablation on AntMaze (normalized score $\times 100$, critic best-of-32). 1-step fails everywhere; 2-step is best or tied; more steps do not help.}
    \label{tab:abl_nfe}
    \begin{tabular}{lcccc}
        \toprule
        \textbf{Maze} & \textbf{1} & \textbf{2} & \textbf{4} & \textbf{8} \\
        \midrule
        medium-diverse & $0.0$ & $\mathbf{80.6}$ & $73.5$ & $79.1$ \\
        large-diverse  & $0.0$ & $\mathbf{49.4}$ & $47.5$ & $42.2$ \\
        large-play     & $0.0$ & $\mathbf{57.8}$ & $38.8$ & $42.6$ \\
        medium-play    & $0.0$ & $74.4$ & $72.6$ & $\mathbf{76.4}$ \\
        \bottomrule
    \end{tabular}
\end{table}

\subsection{Mechanism: Why 1-Step Collapses}
Table~\ref{tab:abl_diag} measures the cross-sample standard deviation over $K$ draws at each NFE. At 1-step the samples form a tight cluster (std $0.044$) far below the dataset action spread ($\approx 0.77$): the draws collapse onto a single, mode-averaged estimate, which is why critic selection over 32 of them still scores $0.0$, there is no valid mode to select. Moving to 2-step roughly doubles the spread ($0.081$), restoring the diversity that makes best-of-$K$ effective. Beyond 2-step the spread saturates, consistent with the performance plateau above.

\begin{table}[ht]
    \centering
    \caption{Cross-sample action standard deviation over $K$ draws as a function of NFE. The dataset action standard deviation is $\approx 0.77$.}
    \label{tab:abl_diag}
    \begin{tabular}{lccc}
        \toprule
        \textbf{NFE} & \textbf{1} & \textbf{2} & \textbf{4} \\
        \midrule
        cross-sample std & $0.044$ & $0.081$ & $0.086$ \\
        \bottomrule
    \end{tabular}
\end{table}

\subsection{Distillation Time Grid}
The consistency loss is evaluated on adjacent pairs of a time grid, and both its spacing and density affect distillation quality. Table~\ref{tab:abl_grid} reports the effect on halfcheetah-medium-expert. Spacing the grid evenly in log-SNR rather than evenly in $t$ substantially improves the distilled student ($72.4$ vs $49.4$ at $N{=}20$): a log-SNR grid equalizes the per-step teacher-ODE accuracy, whereas a linear grid leaves the largest, least accurate steps concentrated at high noise. Grid density shows the opposite of a ``more is better'' trend: increasing the node count from $N{=}20$ to $N{=}80$ degrades performance ($72.4$ to $46.0$), as a denser grid shortens each consistency jump but accumulates per-step error over more steps. We therefore use a log-SNR grid with $N{=}20$. These results are single-seed.

\begin{table}[ht]
    \centering
    \caption{Distillation-grid ablation on halfcheetah-medium-expert (normalized score $\times 100$, single-seed). A log-SNR grid at $N{=}20$ is best on both axes.}
    \label{tab:abl_grid}
    \begin{tabular}{lcc}
        \toprule
        \textbf{Axis} & \textbf{Setting} & \textbf{Score} \\
        \midrule
        \multirow{2}{*}{Spacing ($N{=}20$)} & linear   & $49.4$ \\
                                            & log-SNR  & $\mathbf{72.4}$ \\
        \midrule
        \multirow{2}{*}{Density (log-SNR)}  & $N{=}20$ & $\mathbf{72.4}$ \\
                                            & $N{=}80$ & $46.0$ \\
        \bottomrule
    \end{tabular}
\end{table}

\subsection{Summary}
Together these ablations establish that both upgrades over deterministic single-action extraction are load-bearing: the critic ranking converts drawn candidates into a large performance gain (Table~\ref{tab:abl_selection}), and the 2-step sampler is required for those candidates to be valid in the first place (Tables~\ref{tab:abl_nfe},~\ref{tab:abl_diag}). The nuPlan best-of-$K$ sweep in the main paper confirms the same saturation behavior in closed-loop driving.

\section{Continuous-Time Student (sCM): The Non-Deployed Sibling}
\label{sec:scm}
This Section complements Section IV of the main paper and provides the evidence for the reward-guidance dichotomy reported there. Alongside the deployed discrete student, we develop a continuous-time consistency student based on TrigFlow~\cite{lu2025simplifying}. We present its formulation and results here, and show that while it excels in the unimodal-precision regime, reward guidance destabilizes it on multimodal navigation, which is why the discrete student is deployed for driving.

\subsection{Formulation}
The continuous-time student uses the TrigFlow parameterization with data scale $\sigma_d{=}1$ and prior $\mathcal{N}(0,I)$ at $t{=}T_{\max}$. The forward process is
\begin{equation}
x_t = \cos(t)\,a_0 + \sin(t)\,z, \qquad z\sim\mathcal{N}(0,I),
\end{equation}
and the consistency function is parameterized through a raw velocity network $F_\theta$,
\begin{equation}
f_\theta(x_t,t,s) = \cos(t)\,x_t - \sin(t)\,F_\theta(x_t,t,s),
\end{equation}
which satisfies the boundary condition $f_\theta(x,0)=x$ exactly. No EDM preconditioning, tanh, or dropout is used.

\noindent\textbf{Teacher wrapping.} The frozen VP teacher is reused by mapping TrigFlow time $t$ to VP time $\tau=\phi(t)$ via $\alpha_\tau=\cos(t)$. Solving the VP log-mean-coefficient relation gives the positive root
\begin{equation}
\tau = \frac{-b+\sqrt{b^2-4aL}}{2a}, \quad
a=\tfrac{1}{4}(\beta_1-\beta_0),\; b=\tfrac{1}{2}\beta_0,\; L=\log\cos(t),
\end{equation}
so that $(\alpha_\tau,\sigma_\tau)=(\cos t,\sin t)$ and the VP-noised state coincides with the TrigFlow state. The PF-ODE velocity from the teacher (stop-gradient) is
\begin{equation}
v_t = \frac{\epsilon_\psi(x_t,\phi(t),s) - \sin(t)\,x_t}{\cos(t)} = \cos(t)\,z - \sin(t)\,a_0.
\end{equation}

\noindent\textbf{Continuous consistency loss.} The tangent $\tfrac{df_\theta}{dt}$ is obtained by a forward-mode JVP of $f_\theta$ along $(x_t,t)$ with tangent $(v_t,1)$, and $F_\theta$ is pushed by the normalized tangent (sCM tangent normalization),
\begin{equation}
g = \cos(t)\,\frac{df_\theta}{dt}, \quad
\mathcal{L}_{\text{sCM}} = \mathbb{E}\Big\lVert F_\theta - \mathrm{sg}\!\big[F_\theta + \tfrac{g}{\lVert g\rVert + c}\big]\Big\rVert^2,
\end{equation}
with constant $c$. The raw $\lVert g\rVert$ is the consistency residual, logged and driven toward zero.

\noindent\textbf{Velocity anchor.} $F_\theta$ is jointly regressed to the true flow velocity at all $t$,
\begin{equation}
v_{\text{target}} = \cos(t)\,z - \sin(t)\,a_0, \quad
\mathcal{L}_{\text{diff}} = \mathbb{E}\lVert F_\theta - v_{\text{target}}\rVert^2,
\end{equation}
providing the coherent velocity field the consistency term sharpens. Times are drawn from $\log\tan(t)\sim\mathcal{N}(p_{\text{mean}},p_{\text{std}})$. The full objective is $\mathcal{L}=w_q\mathcal{L}_Q + w_{\text{cons}}\mathcal{L}_{\text{sCM}} + w_{\text{diff}}\mathcal{L}_{\text{diff}}$, with $\mathcal{L}_Q$ structured as in the discrete student. Unlike the discrete student, there is \emph{no EMA target, no discrete time grid, and no low-noise data anchor}; these are the structural differences that the dichotomy below depends on.

\subsection{The Reward-Guidance Dichotomy}
\textbf{Reward-free, sCM is stable and strong.} Without the reward term, the continuous student excels on the unimodal-precision regime, exceeding the discrete student on the expert locomotion tasks (Table~\ref{tab:scm_noq}), and is even competitive on AntMaze.

\begin{table}[ht]
    \centering
    \caption{Reward-free continuous student (scm-noq), best-of-$K$ ($K{=}32$), normalized score $\times 100$ (mean $\pm$ std).}
    \label{tab:scm_noq}
    \begin{tabular}{lc}
        \toprule
        \textbf{Task} & \textbf{sCM (reward-free)} \\
        \midrule
        walker2d-medium-expert     & $111.6 \pm 0.0$ \\
        halfcheetah-medium-expert  & $91.0 \pm 0.9$ \\
        hopper-medium              & $65.2 \pm 1.2$ \\
        walker2d-medium            & $85.5 \pm 0.9$ \\
        hopper-medium-replay       & $90.6 \pm 1.3$ \\
        walker2d-medium-replay     & $88.4 \pm 0.4$ \\
        \midrule
        antmaze-umaze              & $86.0 \pm 1.6$ \\
        antmaze-medium-diverse     & $82.3 \pm 2.1$ \\
        antmaze-medium-play        & $74.3 \pm 2.6$ \\
        \bottomrule
    \end{tabular}
\end{table}

\noindent\textbf{Reward-on, sCM collapses on multimodal navigation.} Activating the reward term collapses the continuous student on AntMaze: medium-diverse falls from $82.3$ to $2.0$, and the large mazes drop to $0.0$ (Table~\ref{tab:scm_q}). The collapse persists across reward weights ($w_q{=}3$ on medium-diverse; $w_q{\in}\{0.5,1.0\}$ on the large mazes), so it is not a single mis-set hyperparameter. The continuous student has no working reward integration in the multimodal regime.

\begin{table}[ht]
    \centering
    \caption{Reward-on continuous student (scm $+\,Q$) on AntMaze, $K{=}32$, normalized score $\times 100$. Contrast with the reward-free values in Table~\ref{tab:scm_noq}.}
    \label{tab:scm_q}
    \begin{tabular}{lcc}
        \toprule
        \textbf{Task} & \textbf{$w_q$} & \textbf{sCM $+\,Q$} \\
        \midrule
        medium-diverse & $3.0$ & $2.0$ \\
        large-diverse  & $0.5$ & $0.0$ \\
        large-diverse  & $1.0$ & $0.0$ \\
        large-play     & $0.5$ & $0.0$ \\
        large-play     & $1.0$ & $0.0$ \\
        \bottomrule
    \end{tabular}
\end{table}

\subsection{Late-Training Stability by Regime}
The dichotomy also appears in training dynamics on high-precision locomotion. On halfcheetah-medium-expert, the discrete student peaks early and then degrades over training: return rises to roughly $45$ by epoch~38 before falling to about $23$ by epoch~100, while its consistency loss drifts upward ($0.11\to0.27$). The continuous student on the same task is stable: return climbs to about $91$ and holds, with a flat consistency loss ($\approx0.14$). This is consistent with the dichotomy, as the continuous student's all-$t$ velocity anchor maintains a coherent field on the single near-optimal mode, whereas the discrete student's low-noise-only anchor does not fully constrain the high-noise consistency bootstrap in this precision regime. We note that this comparison is between two methods on the same task and is not a controlled isolation of any single component (the runs differ in more than one factor); it is reported as evidence for the regime split, not as a causal ablation of a single mechanism. This result is single-seed.

\subsection{Discussion}
The two students are structural mirror images. The discrete student absorbs reward guidance on multimodal navigation, anchored by its low-noise denoising anchor, but trails on unimodal-precision locomotion. The continuous student is the opposite: stable and accurate in reward-free precision control, but lacking this data anchor, it has no anchor to keep it on the data manifold under reward guidance and collapses once reward guidance is applied to multimodal navigation. This dichotomy is why the discrete student is the one deployed for driving, where multimodal closed-loop reactivity under a safety reward is exactly the demanding regime. We note one honest limitation of this evidence: the reward-on collapse is demonstrated on AntMaze, the multimodal proxy, and we do not have reward-on continuous-student results on locomotion; the claim is therefore that reward guidance destabilizes the continuous student \emph{on multimodal navigation}, not universally.

\section{Additional Qualitative Analysis}
This Section compliments Section IV-B of the main paper.
\subsection{Inference Timing and Decision-Making Comparison}
\begin{figure*}[ht]
    \centering
    % --- Top: Teacher Model Failure ---
    \subfloat{
        \begin{minipage}{\textwidth}
            \centering
            \includegraphics[width=0.70\linewidth]{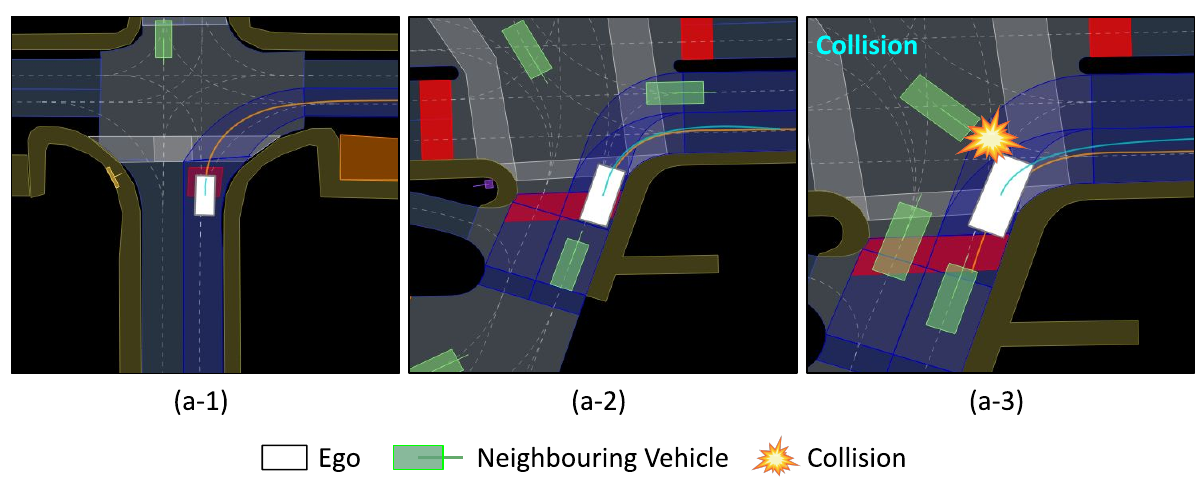}\\[0.5ex]
            \footnotesize (a) \textbf{DiffusionPlanner}: Collision occurs at Frame 35 following a later turn commitment.
            \label{fig:diffusion_failed}
        \end{minipage}
    }

    \vspace{1.5em} % Clear separation between models

    % --- Bottom: Distilled Policy Success ---
    \subfloat{
        \begin{minipage}{\textwidth}
            \centering
            \includegraphics[width=0.70\linewidth]{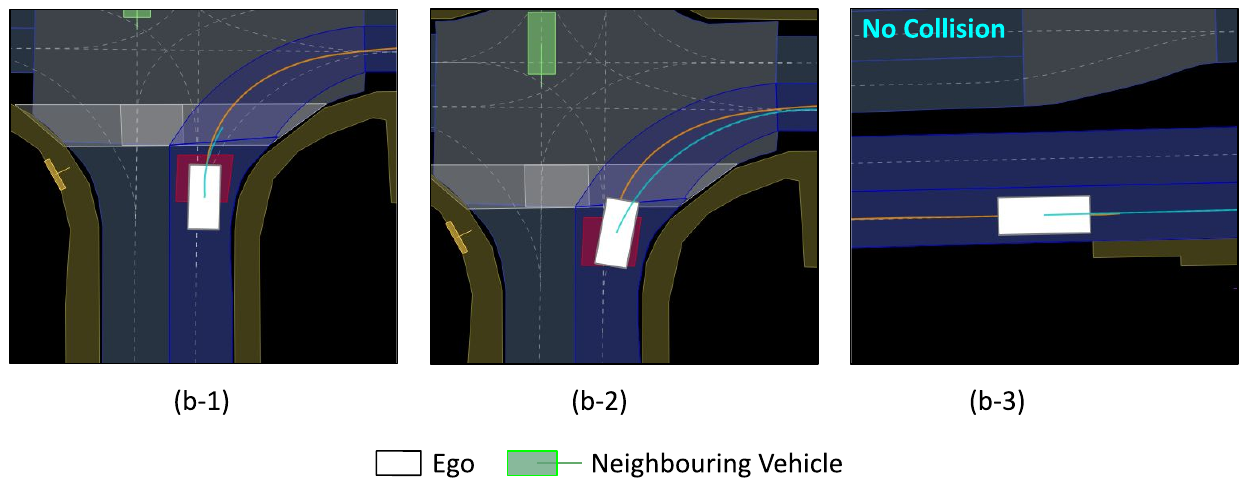}\\[0.5ex]
            \footnotesize (b) \textbf{RAPiD (Ours)}: Safe maneuver completed by Frame 20; the higher PDM safety and proximity weighting commits to the turn earlier, preserving clearance from the moving cross-traffic.
            \label{fig:our_method_success}
        \end{minipage}
    }

    \caption{Comparison of real-time decision making in a right-turn edge case from nuPlan's ``making right turn'' scenario. The distilled policy (bottom) commits to the turn 15 frames earlier than
the DiffusionPlanner (top) and successfully averts the collision.}
\end{figure*}

A direct comparison of the planning timelines reveals that RAPiD initiates safety maneuvers much earlier than the teacher model. As shown in Figure~\ref{fig:our_method_success}(b-1), RAPiD has already initiated a right-turn trajectory, demonstrating its ability to plan ahead for the maneuver. In contrast, DiffusionPlanner~\cite{zheng2025diffusion} in Figure~\ref{fig:diffusion_failed}(a-1) shows no indication of planning for the turn. In Figure~\ref{fig:diffusion_failed}(a-2), the teacher model finally initiates a right-turn decision. However, this is a delayed decision that occurs too late to safely account for the surrounding traffic. As seen in Figure~\ref{fig:diffusion_failed}(a-3), this later turn commitment is followed by a collision. The earlier commitment is consistent with the combined effects of lower inference latency and PDM-guided trajectory selection. By increasing the Comfort weight to 5 (from the nuPlan~\cite{karnchanachari2024towards} weight of 2) and applying a Proximity multiplier of 5, the policy becomes more sensitive to hazardous spatial relationships. While the teacher model focuses on Progress Along Route (nuPlan weight: 5), the distilled policy prioritizes safety and comfort, allowing it to act to successfully avert the collision.

\subsection{interPlan Scenario Based Analysis}
\begin{figure*}[ht]
    \centering
    \includegraphics[width=1.0\linewidth]{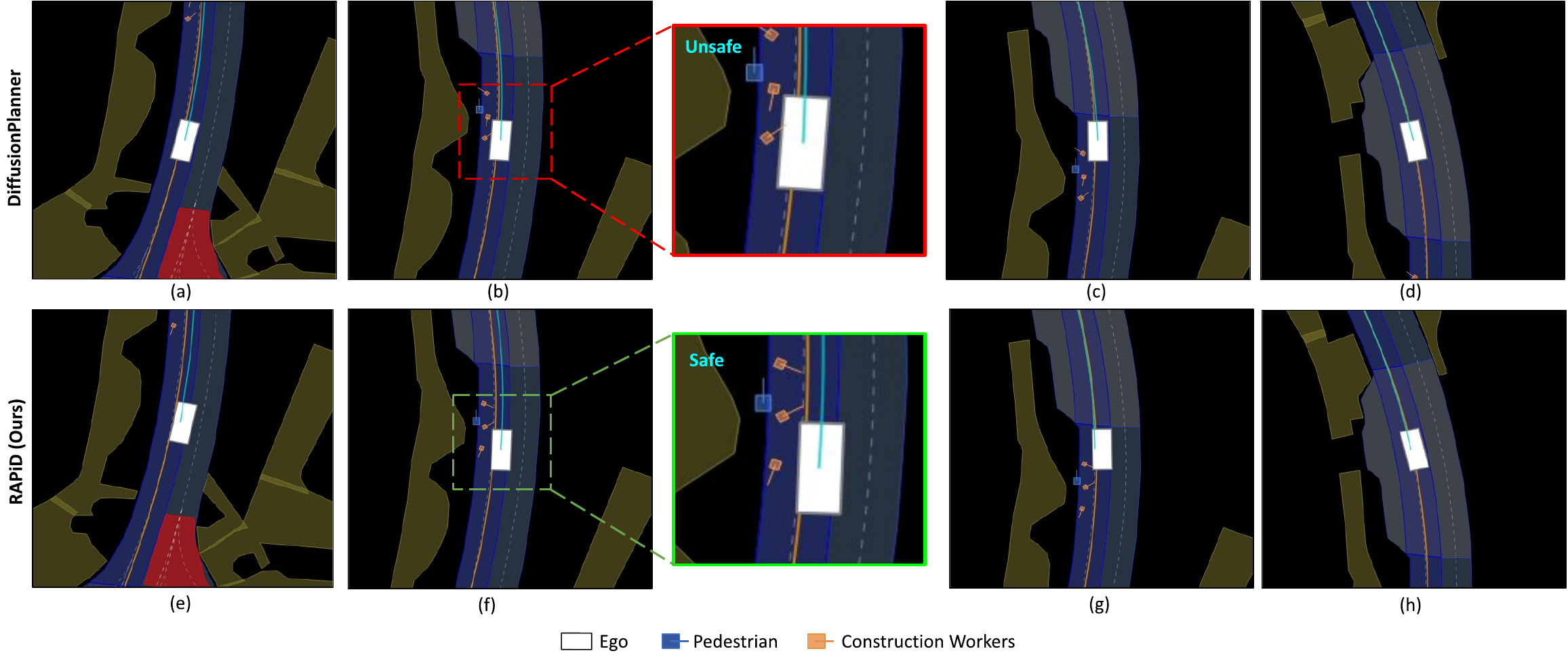}
    \caption{Construction zone scenario comparison between DiffusionPlanner (a-d) and RAPiD (e-h) in interPlan}
    \label{constr}
\end{figure*}

\begin{figure*}[ht]
    \centering
    \includegraphics[width=1.0\linewidth]{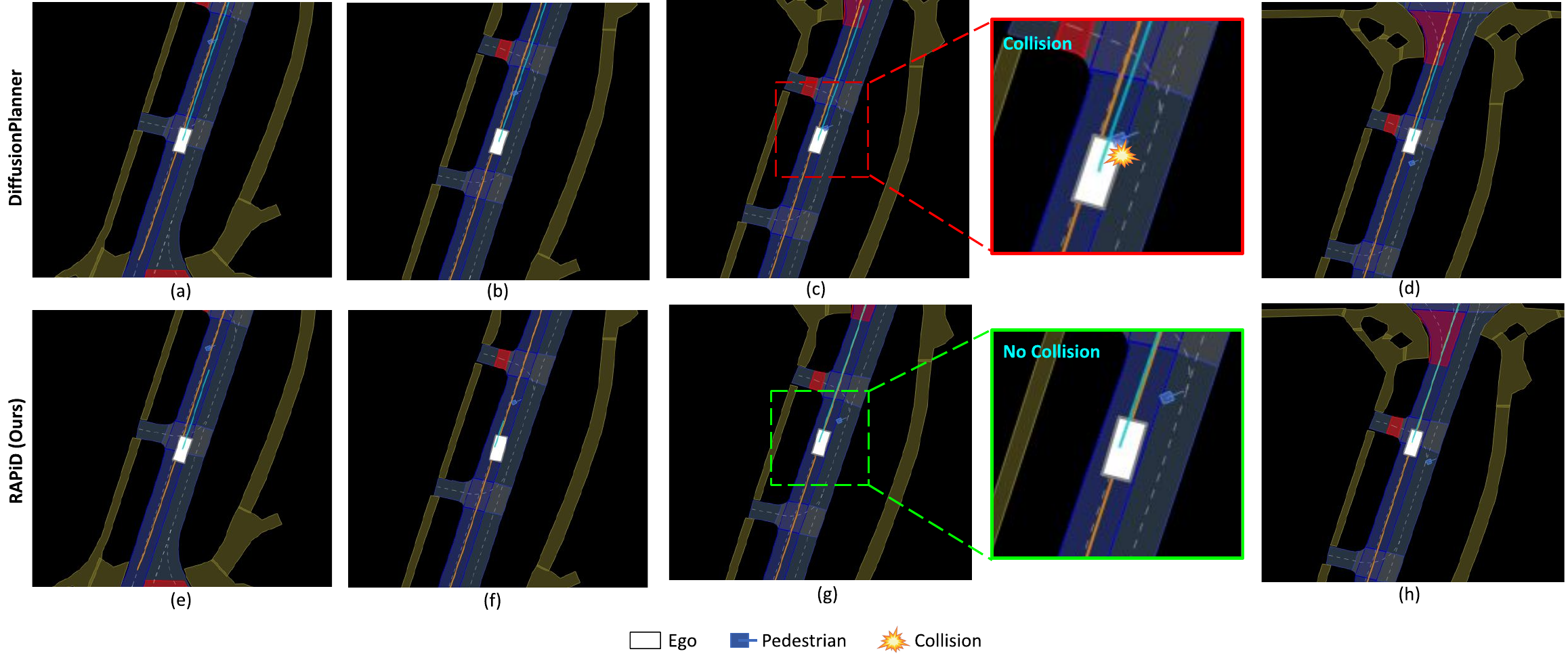}
    \caption{Jaywalking pedestrian scenario comparison between DiffusionPlanner (a-d) and RAPiD (e-h) in interPlan}
    \label{jayw}
\end{figure*}

\begin{figure*}[ht]
    \centering
    \includegraphics[width=1.0\linewidth]{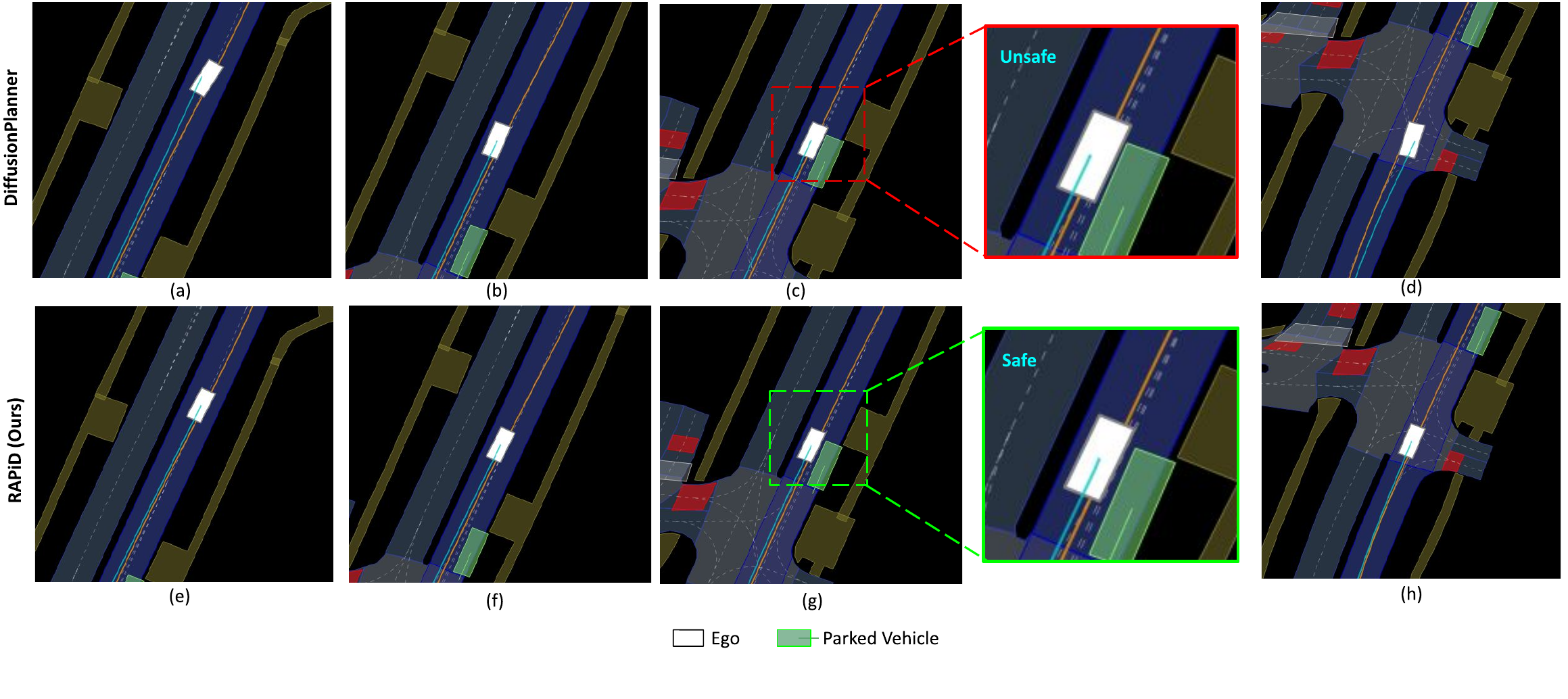}
    \caption{Nudge scenario with partial lane obstruction. DiffusionPlanner (a-d) vs RAPiD (e-h) in interPlan}
    \label{nudge}
\end{figure*}

\begin{figure*}[ht]
    \centering
    \includegraphics[width=1.0\linewidth]{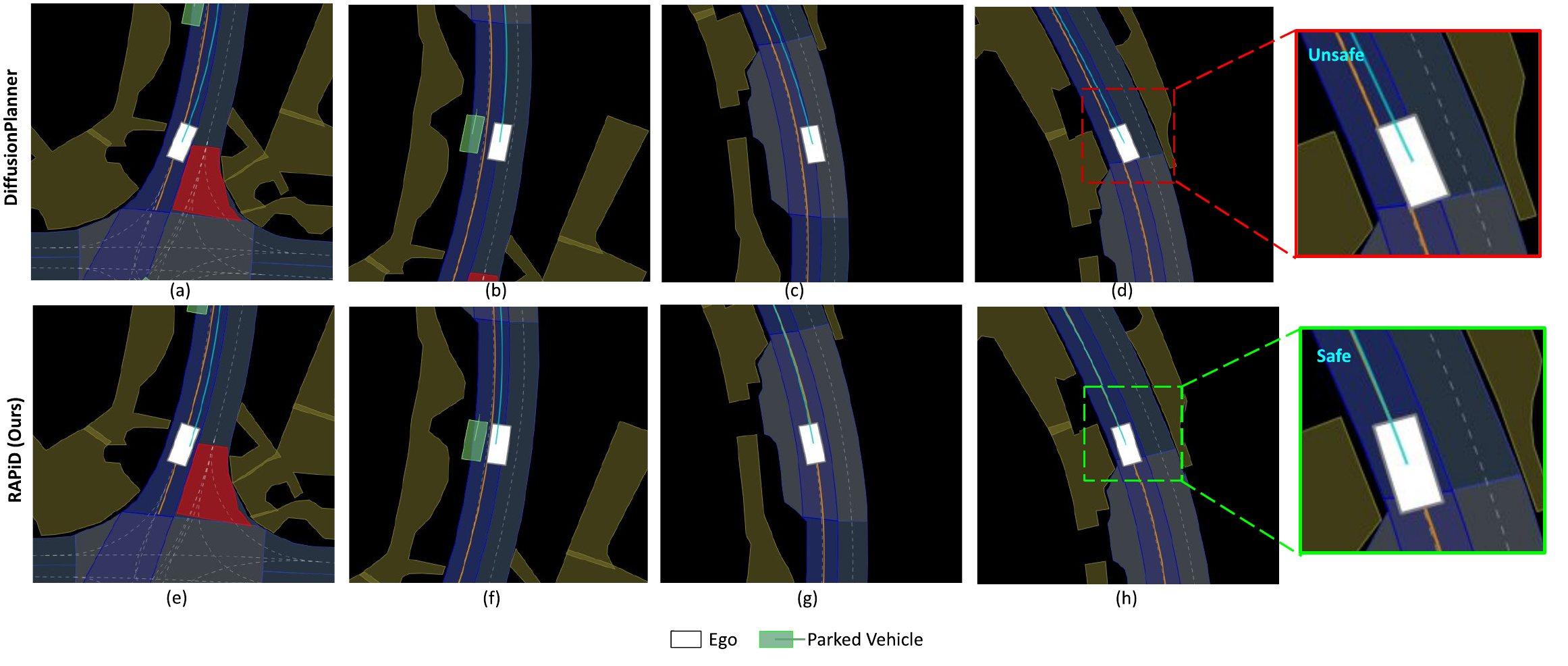}
    \caption{Overtake scenario with parked vehicle. DiffusionPlanner (a-d) vs RAPiD (e-h) in interPlan}
    \label{overtk}
\end{figure*}

We present detailed qualitative analysis of four challenging scenarios from the interPlan~\cite{hallgarten2023prediction} benchmark. Figures~\ref{constr}-\ref{overtk} show representative frames where DiffusionPlanner (frames a-d) and RAPiD (frames e-h) exhibit different planning behaviors. The predicted trajectories are shown as overlays on the scene images. These scenarios were selected to demonstrate critical safety differences in handling construction zones, jaywalking pedestrians, lane obstructions (nudge), and overtaking maneuvers. \newline
\textbf{Construction Zone:} In Figure~\ref{constr}(a), DiffusionPlanner begins moving away from the centerline toward the oncoming lane to navigate around the construction zone and workers. However, in Figure~\ref{constr}(b), we can see that despite this lateral movement, DiffusionPlanner does not maintain enough distance and comes dangerously close to the construction workers. In comparison, RAPiD begins its lateral maneuver earlier as shown in Figure~\ref{constr}(e), moving toward the oncoming lane while keeping a larger buffer from the construction zone. Figure~\ref{constr}(f) shows RAPiD successfully balancing between maintaining safe distance from the workers while not moving too far into the oncoming lane. Both planners return to the centerline after passing the construction zone, as seen in Figure~\ref{constr}(d) for DiffusionPlanner and Figure~\ref{constr}(h) for RAPiD. \newline
\textbf{Jaywalking:} Figure~\ref{jayw}(a) shows a pedestrian crossing the road mid-street (jaywalking), but DiffusionPlanner plans a trajectory that goes straight through where the pedestrian is standing. In Figure~\ref{jayw}(b-c), we see that DiffusionPlanner does not adjust its plan in time and ends up colliding with the pedestrian. This represents a failure to prioritize safety over continuing along the route. In contrast, Figure~\ref{jayw}(e) shows RAPiD planning to stop before reaching the pedestrian's position. Figure~\ref{jayw}(f) shows RAPiD has come to a near-complete stop to avoid hitting the pedestrian. Once the pedestrian has moved out of the lane in Figure~\ref{jayw}(g), RAPiD resumes driving in its intended direction as shown in Figure~\ref{jayw}(h). \newline
\textbf{Nudge:} In the nudge scenario, a vehicle is blocking part of the lane. Figure~\ref{nudge}(a) shows DiffusionPlanner making a sharp lateral movement away from the blocked portion of the lane. However, Figure~\ref{nudge}(b) reveals that this distance is still insufficient, resulting in a near-collision where the ego vehicle almost side-swipes the parked vehicle. RAPiD's approach in Figure~\ref{nudge}(e) shows a smoother, more gradual lateral movement that begins earlier. Figure~\ref{nudge}(f) demonstrates that RAPiD maintains a larger safety margin from the obstruction compared to DiffusionPlanner. This smoother trajectory planning allows RAPiD to avoid the near-collision situation entirely, as shown in Figure~\ref{nudge}(g). Additionally, RAPiD returns to the lane center more smoothly in Figure~\ref{nudge}(h), whereas DiffusionPlanner's return in Figure~\ref{nudge}(d) appears more abrupt. \newline
\textbf{Overtake:} In Figure~\ref{overtk}(a), DiffusionPlanner detects a parked vehicle blocking part of its lane and begins moving into the oncoming lane to pass it. Figure~\ref{overtk}(b) shows DiffusionPlanner successfully maintaining distance from the parked vehicle during the overtake. However, Figure~\ref{overtk}(c-d) reveals a problem: DiffusionPlanner continues drifting into the oncoming lane even after passing the obstacle and fails to return to its own lane. This creates a dangerous situation with high risk of head-on collision. RAPiD's behavior in Figure~\ref{overtk}(e) shows the beginning of its overtake maneuver, and Figure~\ref{overtk}(f) demonstrates that RAPiD maintains a safe distance from the parked vehicle while minimizing how far it moves into the oncoming lane. Most importantly, Figure~\ref{overtk}(g) shows that RAPiD has already returned to the center of its own lane immediately after passing the obstacle, completing the maneuver safely.

%\newpage
\bibliographystyle{IEEEtran}
\bibliography{ref}

% \end{document} % This now contains the \clearpage at the end

\end{document}